\documentclass[default,iicol]{sn-jnl}

\usepackage{colortbl}
\usepackage{multirow}
\usepackage{xcolor}
\usepackage{rotating}
\usepackage{bm}
\usepackage{mathtools}
\usepackage{amsmath, amsfonts}
\usepackage[inline]{enumitem}
\usepackage{bbding}
\usepackage[switch]{lineno}  
\usepackage{booktabs}

\usepackage{pict2e}
\usepackage[calc]{picture}

\newcommand{\bx}{\ensuremath{\bm{x}}}

\newcommand{\bz}{\ensuremath{\bm{z}}}

\newcommand{\be}{\ensuremath{\bm{e}}}
\newcommand{\bp}{\ensuremath{\bm{p}}}
\newcommand{\ba}{\ensuremath{\bm{a}}}
\newcommand{\bh}{\ensuremath{\bm{h}}}

\newcommand{\bbr}{\ensuremath{\bm{r}}}   

\newcommand{\bs}{\ensuremath{\bm{s}}}

\newcommand{\bZ}{\ensuremath{\bm{Z}}}

\newcommand{\bE}{\ensuremath{\bm{E}}}

\newcommand{\bF}{\ensuremath{\bm{F}}}
\newcommand{\bA}{\ensuremath{\bm{A}}}

\newcommand{\bM}{\ensuremath{\bm{M}}}
\newcommand{\bI}{\ensuremath{\bm{I}}}
\newcommand{\bmu}{\ensuremath{\bm{\mu}}}
\newcommand{\bsigma}{\ensuremath{\bm{\sigma}}}
\newcommand{\bphi}{\ensuremath{\bm{\phi}}}
\newcommand{\balpha}{\ensuremath{\bm{\alpha}}}

\definecolor{colorblue}{rgb}{0.29, 0.59, 0.82}
\definecolor{colorred}{rgb}{0.9, 0.17, 0.31}
\definecolor{colorred}{rgb}{0.9, 0.17, 0.31}
\definecolor{coolgrey}{rgb}{0.55, 0.57, 0.67}
\definecolor{colorwhite}{rgb}{1, 1, 1}
\newcommand{\dcb}{\cellcolor{colorwhite}}

\newcommand{\mr}[1]{\multirow{2}{*}{#1}}
\newcommand{\mrm}[1]{\multirow{-2}{*}{#1}}

\newcolumntype{"}{@{\hskip\tabcolsep\vrule width 1.5pt\hskip\tabcolsep}}
\makeatletter
\newcommand{\thickhline}{%
    \noalign {\ifnum 0=`}\fi \hrule height 1pt
    \futurelet \reserved@a \@xhline
}

\newcommand{\fun}[1]{\ensuremath{\mathit{#1}}}
\newcommand{\func}[2]{\ensuremath{\fun{#1}\left(#2\right)}}

\let\oldSum\sum

\jyear{2021}

\begin{document}

\title[Article Title]{RGCVAE: Relational Graph Conditioned Variational Autoencoder for Molecule Design}

\author*[1,2]{\fnm{Davide} \sur{Rigoni}}\email{davide.rigoni.2@phd.unipd.it}

\author[1]{\fnm{Nicolo'} \sur{Navarin}}\email{nnavarin@math.unipd.it}

\author[1]{\fnm{Alessandro} \sur{Sperduti}}\email{sperduti@math.unipd.it}
        
\affil*[1]{\orgdiv{Department of Mathematics}, \orgname{University of Padua}, \orgaddress{\street{Via Trieste 63}, \city{Padua}, \postcode{35131}, \state{Padova}, \country{Italy}}}
\affil[2]{\orgdiv{DKM}, \orgname{Fondazione Bruno Kessler}, \orgaddress{\street{Via Sommarive, 18}, \city{Povo}, \postcode{38123}, \state{Trento}, \country{Italy}}}

\abstract{
Identifying molecules that exhibit some pre-specified properties is a difficult problem to solve.
In the last few years, deep generative models have been used for molecule generation.
Deep Graph Variational Autoencoders are among the most powerful machine learning tools with which it is possible to address this problem. 
However, existing methods struggle in capturing the true data distribution and tend to be computationally expensive.

In this work, we propose RGCVAE, an efficient and effective Graph Variational Autoencoder based on: \textit{(i)} an encoding network exploiting a new powerful Relational Graph Isomorphism Network; \textit{(ii)} a novel probabilistic decoding component.
Compared to several state-of-the-art VAE methods on two widely adopted datasets, RGCVAE shows state-of-the-art molecule generation performance while being significantly faster to train.
}


\maketitle
\section*{Acknowledgments}
The authors acknowledge the HPC resources of the Department of Mathematics, University of Padua, made available for conducting the research reported in this paper. This research was partly supported by the SID/BIRD 2020 project "Deep Graph Memory Networks", University of Padua.

\section{Introduction}\label{sec:introduction}

Chemical space is huge, and it contains many (unknown) molecules which can turn useful in many industrial and pharmaceutical areas, e.g. to develop new improved drugs. The problem is how to search such space in an efficient and targeted way. 
Traditional approaches, such as high throughput screening~\cite{curtarolo2013high, pyzer2015high}, combinatorial and evolutionary algorithms~\cite{devi2015evolutionary}, are substantially based on brute force search, possibly improved using heuristics. More recent approaches are based on availability of data (e.g., a corpus of molecules with known properties) to which Machine Learning approaches are applied.
For example, Recursive Neural Networks or graph kernels have been used to predict properties of pre-defined or commercial compounds \cite{QSPRAR_NNS, DBLP:journals/jcisd/BernazzaniDMMSST06}, and Generative Models have been used to generate candidate molecules that are likely to exhibit some pre-specified properties~\cite{Blaschke2018,Oglic2018}.

Generation of candidate molecules seems to be a quite promising approach since, based
on a family of compounds for which the property of interest is known, the model should be able to capture the common features that deliver the property of interest. If this is the case, the generation process is also required to produce molecules which are novel and useful.

Models for the generation of new molecules mainly represent the molecule structures as:
\begin{enumerate*}[label=(\roman*)]
    \item SMILES strings that describe the molecule structures, such as in~\cite{weininger1988smiles, weininger1989smiles, weininger1990smiles};
    \item graph representations, in which nodes represent atoms and edges represent bonds, such as in~\cite{rigoni2020conditional, DBLP:conf/nips/LiuABG18, DBLP:conf/nips/MaCX18};
    \item graph representations, in which nodes represents molecules' substructures, such as in~\cite{DBLP:conf/icml/JinBJ18, jin2020hierarchical}.
\end{enumerate*}
Since it is simpler to deal with strings than graphs, early works adopted the SMILES representation, while more recent works use the graph representation because of its greater expressiveness.
Both the SMILES representation and the graph representation of molecules have advantages and disadvantages.
For example, graph representations are more expressive than the SMILES representations, but in contrast, are harder to learn.

Several early works are based on the Variational Autoencoder (VAE)~\cite{diederik2014auto} approach, such as Character VAE~\cite{gomez2018automatic}  which uses a variational autoencoder to directly predict the molecule SMILES representation, Grammar VAE~\cite{DBLP:conf/icml/KusnerPH17} which adds a \emph{context-free grammar} to Character VAE to guide the correct generation of SMILES strings, and Syntax Directed VAE~\cite{DBLP:conf/iclr/DaiTDSS18} which adds a more expressive grammar, the \emph{attribute grammar}~\cite{knuth1968semantics}, capable of generating a higher number of valid SMILES strings.
The VAE approach has also been used by models which consider the graph representation of the molecules, e.g. Junction Tree VAE~\cite{DBLP:conf/icml/JinBJ18}, Regularized Graph VAE~\cite{DBLP:conf/nips/MaCX18}, Constrained Graph VAE (CGVAE)~\cite{DBLP:conf/nips/LiuABG18}, and Conditional CGVAE~\cite{rigoni2020conditional}.

Other works are based on recurrent neural networks~\cite{segler2018generating, podda2020deep}, adversarial autoencoders~\cite{DBLP:journals/corr/MakhzaniSJG15,kadurin2017drugan, polykovskiy2018entangled}, generative adversarial networks~\cite{arjovsky2017wasserstein,guimaraes2017objective} and Flow based models \cite{DBLP:conf/iclr/ShiXZZZT20, DBLP:conf/kdd/ZangW20, DBLP:conf/aaai/KuznetsovP21}.
More recently, some works address the generation of new molecules starting from a set of reactants that need to be combined together to form a new molecule \cite{DBLP:conf/nips/BradshawPKSH19, DBLP:conf/nips/BradshawPKSH20}.
To evaluate and compare this variety of models, a systematic model comparison~\cite{rigoni2020systematic} and some frameworks have been developed, such as MOSES~\cite{polykovskiy2018molecular} and GuacaMol~\cite{brown2019guacamol}.

Some of these models, in addition to generating new molecules, allow to optimize a given molecule structure towards those that exhibit a better pre-specified property using:
\begin{enumerate*}[label=(\roman*)]
    \item an ad-hoc optimization process~ \cite{gomez2018automatic, DBLP:conf/icml/KusnerPH17, DBLP:conf/iclr/DaiTDSS18, DBLP:conf/icml/JinBJ18, DBLP:conf/nips/LiuABG18}, or 
    \item introducing additional terms in the loss function~\cite{de2018molgan, DBLP:conf/nips/MaCX18}.
\end{enumerate*}

However, in this work we mainly focus on the \emph{distribution learning task}, which is defined as the task of learning the probability distribution that has generated the molecules in the dataset, instead of the \emph{structure optimization task} that aims to optimize a given molecule structure to exhibit a better pre-specified property.

To solve the \emph{distribution learning task} we focus on the VAE approaches, that seem to deliver the best trade-off between generative capabilities and ease of training. 
Indeed, VAE approaches have been shown to be able to learn distributions even considering different types of data, such as images and text, and to be able to generate new examples that try to follow the same distribution that have generated the examples in the training set.

Existing VAE approaches, however, present two main issues:
\begin{enumerate*}[label=(\roman*)]
\item high reconstruction accuracy of the molecules at the expense of the generation quality and vice versa;
\item low computational efficiency.
\end{enumerate*}
Returning a high reconstruction error for the molecules belonging to the test set is usually accepted in favour of the ability of the model to generate new valid molecules. 
This, however, is an index of poor identification of the input molecule subspace, thus posing doubts on the usefulness of the new generated molecules. 
Generation of valid (and different) molecules is of course a desirable property, however recently developed models, based on deep learning, spend a significant amount of computation time to guarantee it.
In this paper, we address these two issues by proposing a novel computationally efficient VAE-based model that is able to get a low reconstruction error jointly with a high validity index.

\section{Background and Definition}
In this section, we recall the basic concepts on Variational Autoencoders instantiated to graphs. 
We also outline the basic Graph Neural Network model that we use as a starting point for our relational variant.

In this paper, we use the following notation:
\begin{enumerate*}[label=(\roman*)]
    \item bold upper case symbols for matrices and tensors, e.g. $\bA$;
    \item bold lower case symbols for vectors, e.g. $\ba$;
    \item italics upper case symbols for sets, e.g. $A$;
    \item italics lower case symbols for constant values, e.g. $a$.
    \item the position within a tensor or vector is indicated with numeric subscripts, e.g. $\bA_{i, j}$ with $i,j \in \mathbb{N}^+$;
\end{enumerate*} 
We exploit the graph representation of the molecule structure.
Let $Tr$ be the training set of molecules.
Then, $\Theta_n$ and $\Theta_e$ are respectively the set of the atom types and the set of the edge types extracted from $Tr$.
Each molecule in $Tr$ exhibits some pre-specified properties whose set is represented as $\Theta_p$.
Let $d_n = \mid\Theta_n\mid$ be the number of atom types, $d_e = \mid\Theta_e\mid$ be the number of edge types and $d_p = \mid\Theta_p\mid$ be the number of properties.
A molecule is represented as a graph $G = (V, E, \Phi_e, \Phi_n)$ in which $V=\{v_1, \ldots, v_m\}$ is the set of nodes, $E \subseteq \{(v, u) \mid (v, u) \in V^2, v \neq u\}$ is the set of edges, $\Phi_e$ is the edge labeling function $\Phi_e: E \rightarrow \Theta_e$ and $\Phi_n$ is the node labeling function $\Phi_n: V \rightarrow \Theta_n$.
Given a node $v \in V$, the set of its neighbors is given by the function $\mathcal{N}(v) = \{ u \mid  u \in V,  (v, u) \in E \}$.
Given the set of all the graphs ${\cal G}$, for each property $p \in \Theta_p$, the function $\Phi_p: {\cal G}  \rightarrow \mathbb{R}$  returns its value for the molecule.
Each atom $v \in V$ has a valence number $\nu_v \in \{1,\ldots, 8\}$, where 8 is the maximum atom valence according to the periodic table, defined by its atom type $\Phi_n(v)$.

We can equivalently represent a molecule $G$ by an adjacency matrix $\bA \in \{0,1\}^{m\times m}$, an edge type tensor \mbox{$\bE \in \{0,1\}^{m \times m \times d_e}$}, a node attribute matrix $\bF \in \{0,1\}^{m \times d_n}$, and a vector of the molecule's properties values $\bp \in \mathbb{R}^{d_p}$, 
such that $\forall n_1, n_2  \in V$ with indices $i, j \in \{0, ..., m-1\}$, $\forall e \in \Theta_e$ with index $t\in\{0, d_e-1\}$, $\forall p \in \Theta_p$ with index $k\in\{0, ..., d_p-1\}$:

   $ [\bA_{i,j} =1 \iff (n_1,n_2)\in E]$, $[\bE_{i,j, t} =1 \iff (n_1,n_2)\in E \ \land \ \Phi_e((n_1,n_2)) = e]$, $[\bF_{i,z} =1 \iff  \Phi_n(i) = z]$, $[\bp_{k} = \Phi_p(G)]$.

Given a molecule, we can define the histogram of valences as a vector $\balpha$ where, representing with $\nu$ the maximum atom valence number in the $Tr$, $\balpha[i]$ with $i \in \{1, .., \nu\}$ is the number of atoms with valence equal to $i$.

Finally, given a training set $Tr$, the histograms distribution ${\cal H}$ is the probability distribution obtained considering all the histograms of valences of the molecules in $Tr$.

For the sake of clarity, Table~\ref{tab:notation} reports a summary of the main symbols used in this article.
\begin{table*}
\begin{center}
\begin{minipage}{1\linewidth}
\caption{Notation table.}
\label{tab:notation}
\begin{tabular}{ll}
\toprule
\textbf{Symbol}        &\textbf{Description}                                     \\
\midrule
$Tr$                            &Training set of molecules.                             \\ 
$\Theta_n$                      &Set of atom types.                                 \\
$\Theta_e$                      &Set of edge types.                                 \\
$\Theta_p$                      &Set of molecule properties.                        \\
$d_n$                           &Number of atom types.                                  \\
$d_e$                           &Number of edge types.                                  \\
$d_p$                           &Number of properties.                                  \\
$G = (V, E, \Phi_e, \Phi_n)$    &Graph representation of a molecule.                    \\
$V$                             &Set of nodes.                                          \\
$E$                             &Set of edges.                                          \\
$\Phi_e(\cdot)$                 &Edge labeling function.                                \\
$\Phi_n(\cdot)$                 &Node labeling function.                                \\
$\Phi_p(\cdot)$                 &Property labelling function for each molecule property $p \in \Theta_p$.  \\
${\cal G}$                      &Set of all the graphs.                                 \\
$\mathcal{N}(v)$                &Set of neighbors of node $v \in V$.                    \\
$\nu$                           &Max atom's valence number in $Tr$.                     \\
$\nu_v$                         &Valence number of atom $v \in V$.                      \\
$m$                             &Number of atoms in a molecule.                          \\
$\bF, \tilde{\bF}$              &Node attribute matrix and predicted node attribute matrix.                      \\
$\bA, \tilde{\bA}$              &Adjacency matrix and predicted adjacency matrix.  \\
$\bE, \tilde{\bE}$              &Edge type tensor and predicted edge type tensor.                               \\
$\bp, \tilde{\bp}$              &Vector of molecule's properties and predicted vector of molecule's properties. \\
$\balpha$                       &Histogram of valences.                                 \\
${\cal H}$                      &Histograms distribution.                               \\
$\mathcal{N}(0, \bI)$           &Standard Normal Distribution.                          \\
$\bI$                           &Identity matrix.                                       \\
$\epsilon^{(k)}$                &GIN/RGIN hyper-parameter of layer $k$.                      \\
$\bh_{v}^{(k)}$                 &$k-th$ hidden state for node $v \in V$ computed by the $k-th$ GIN/RGIN layer.                \\
$s_h$                           &Size of the vector $\bh_{v}^{(k)}$.                 \\
$s_{\text{lt}}$                 &Latent space size.                                     \\ 
$\bmu_v$                        &Vector of means of node $v \in V$.                     \\
$\bsigma^2_v$                   &Vector of variances of node $v \in V$.                 \\ 
$\bm{\Sigma}_v$                 &Covariance matrix where all values, except the diagonal, are set to $0$.    \\
$Z=\{\bz_v\}$                   &Set of latent space vectors.                            \\
$\bz_v$                         &Vector of latent space features for node $v \in V$.                \\ 
$\bbr_v$                        &Vector of features for node $v \in V$ returned by the \emph{atom decoding}.   \\
$s_n$                           &Size of vector $\bbr_v$.                 \\
$\tau_v$                        &Atom type for node $v \in V$ returned by the \emph{atom decoding} procedure.    \\
$\bphi_{v, u}$                  &Permutation invariant representation in input to the \emph{edge decoding} procedure. \\
$\bM^e_{v, u}$                  &Mask not allowing for self loops and edge duplication among nodes $v, u \in V$.  \\
$\bM^t_{v, l, u}$               &Mask enforcing chemical laws among nodes $v, u \in V$ considering bond $l \in \Theta_e$. \\
$\odot$                         &Component-wise multiplication.                          \\
$\text{Concat}(\cdot,\cdot)$    &Concatenation function.                                 \\
$we(\cdot )$                    &Function computing the edge weights.            \\
$\sigma( \cdot )$               &Element-wise sigmoid function.                                       \\
$\text{tanh}( \cdot )$          &Element-wise hyperbolic tangent function.                                     \\
\bottomrule
\end{tabular}
\end{minipage}
\end{center}
\end{table*}

\subsection{Variational Autoencoders}
The VAE approach for graphs contemplates three main components: the \emph{encoder}, the \emph{decoder} and the \emph{optimizer}.

The \emph{encoder} learns to encode an input molecule into a latent space of a fixed dimension.
The encoder can be defined to produce a single representation for the whole input graph, or one for each graph node. In the latter case, that we consider in this paper, the graph representation is distributed across the nodes. 
The training procedure aim is to drive the node encodings to be normally distributed, i.e. \( \mathcal{N}\left(0, \bI\right)\), where $\bI$ is the identity matrix. 

The \emph{decoder} generates the atoms and their type starting from the latent representation(s), and then adding the bonds between them.
In some cases, the decoder is designed to be compliant to some chemical constraints, e.g. isolated atoms can be removed.
To generate new molecules, it is possible to sample point(s) from \( \mathcal{N}\left(0, \bI\right)\) and to decode them into a (possibly valid) molecule via the decoder.

VAE can incorporate an \emph{optimizer} component which can be used to drive the generation process towards molecules that exhibit good values of some pre-specified properties. This is achieved including in the training procedure a function that predicts a pre-specied property. 
In the generation phase, it is possible to generate molecules exhibiting good values for the property of interest starting from random points in the latent space and performing gradient ascent/descent with respect to the predicted property values.

\subsection{Graph Neural Networks}
Different graph neural networks have been proposed in literature in the last few years. In this paper, we focus on one instantiation that has been shown to perform pretty well in different tasks.
Graph Isomorphism Network (GIN)~\cite{DBLP:conf/iclr/XuHLJ19} is a deep neural network for graphs based on a message-passing scheme, that is an iterative neighborhood aggregation scheme in which the representation of each node $v$ is updated with the contribution of the information that it receives from its neighbors $\mathcal{N}(v)$.
Given a number of layers $K$, GIN implements the following update function for each layer $k \in \{1, ..., K\}$:
\begin{equation}
\begin{aligned}
\bh_{v}^{(k)}=& \operatorname{MLP}^{(k)}\Biggl( \left(1+\epsilon^{(k)}\right) \odot \bh_{v}^{(k-1)} \\
&+ \oldSum_{u \in \mathcal{N}(v)} \bh_{u}^{(k-1)}\Biggr),
\end{aligned}
\end{equation}
where $\epsilon^{(k)}$ is a positive small constant, $\bh_v^{(k)}$ represents the hidden state associated to node $v$ computed by the $k$-th GIN layer, and $\text{MLP}^{(k)}$ is a multi-layer perceptron. These hidden states are then used by a \emph{readout} neural network to generate the desidered output.
In the next section, we present an extension of GIN that considers different edge types, that constitutes a novel contribution.

\section{Proposed Model}
\label{sec:model}
\begin{figure*}[t]
\centering
\includegraphics[width=1\textwidth]{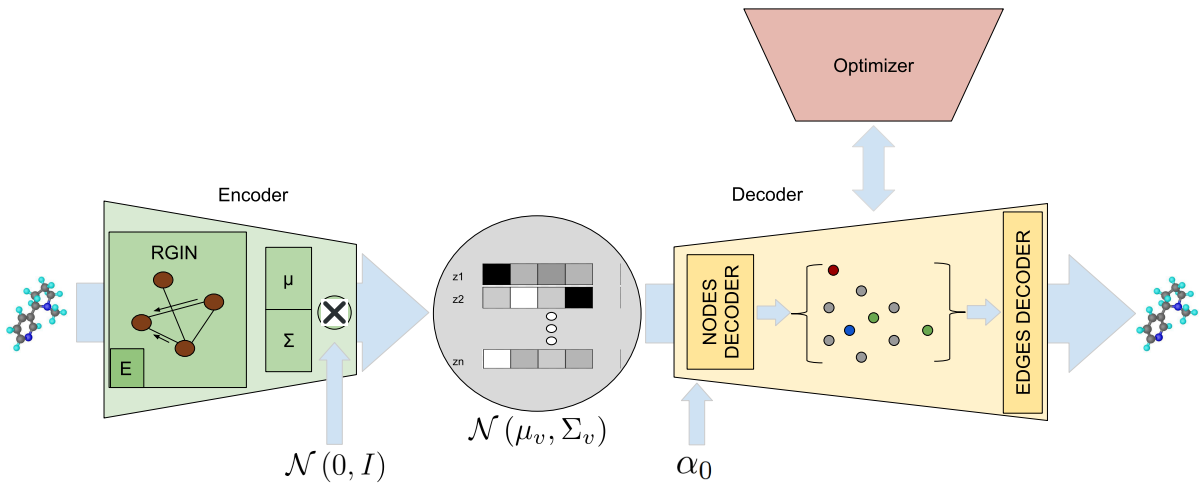}
\caption{RGCVAE model structure. 
The encoder feeds the input molecule to a novel relational version of GIN exploiting an edge-specific neural network $E$ that encodes the molecule in the latent space. The decoder receives in input the sampled points and the initial histogram $\balpha_0$. It decodes the set of sampled points to the unconnected set of atoms, and then it decodes the bonds between them. The optimizer computes an optimized set of latent points which, once decoded, should exhibit better property values.}
\label{fig:RGCVAE}
\end{figure*}

In this section, we present the main components (see Fig.~\ref{fig:RGCVAE}) of the proposed Relational Graph Isomorphism VAE model (RGCVAE)\footnote{Our implementation can be found at: \url{https://github.com/drigoni/RGCVAE}}.

\subsection{Encoder \label{sec:encoder}}
Let us consider the input graph representation $G$  of a molecule with $m$ atoms. The encoder maps each atom $v \in V$ into a vector sampled by a multivariate normal probability distribution with mean \(\bmu_v\) and variance \(\bsigma^2_v\).
To generate these encodings, we define a novel relational variant of GIN, dubbed RGIN, to devise a hidden representation of each atom that embeds the information of its neighboring nodes and of the type of bond connecting them. Our proposed variant is different from the Relational Graph Convolutional Model (R-GCN) presented  in~\cite{schlichtkrull2018modeling}, as will also be clear from the experimental results reported in the ablation study (Section~\ref{sec:ablation}).

In order to consider different types of bonds between atoms, we have changed the original update function of GIN as:
\begin{equation}
\begin{aligned}
    \bh_{v}^{(k)} =
    &\text{MLP}^{(k)}\Biggl(\left(1+\epsilon^{(k)}\right) \bh_{v}^{(k-1)} \\
    &+ \oldSum_{u \in \mathcal{N}(v)} \text{MLP}^{(k)}_{\Phi_e((v,u))}\left(\bh_{u}^{(k-1)}\right)\Biggr),
\end{aligned}
\end{equation}
where $\text{MLP}^{(k)}_{\Phi_e((v,u))}$ is an edge type specific multi-layer perceptron. 
The initial hidden representation state for each \mbox{$v \in V$} is given by $\bh_{v}^{(1)} = W(\Phi_n(v))$, where \mbox{$W:\Theta_n \rightarrow \mathbb{R}^{s_h}$} is a learnable function which returns a feature vector of size $s_h$ for each atom type.
Given a latent space of size $s_{\text{lt}}$, $\forall v \in V$, the encoder maps the above hidden states to the target mean $\bmu_v \in  \mathbb{R}^{s_{\text{lt}}}$ and variance $\bsigma^2_v \in  \mathbb{R}^{s_{\text{lt}}}$ vectors via two different MLPs:
\begin{align}
    \bmu_v &= \text{MLP}_\mu\left(\bh^{(K)}_v\right), \\
    \text{diag}(\bm{\Sigma}_v) &= \bsigma^2_v = \exp \left( \text{MLP}_{\sigma^2}\left(\bh^{(K)}_v\right) \right),
\end{align}
where $\bm{\Sigma}_v$ defines the covariance matrix where all values, except the diagonal, are set to $0$.
It should be emphasized that this model, similarly to CGVAE~\cite{DBLP:conf/nips/LiuABG18}, generates a probability distribution for each node in the graph, unlike other approaches~\cite{gomez2018automatic, DBLP:conf/icml/KusnerPH17, DBLP:conf/iclr/DaiTDSS18, DBLP:conf/icml/JinBJ18, DBLP:conf/nips/MaCX18, de2018molgan} that generate a probability distribution per molecule.

\subsection{Decoder\label{sec:decoder}}
The decoder is constituted by two main components (see Fig.~\ref{fig:RGCVAE}).
The first component implements an \emph{atom decoding} procedure which generates a set of initially unconnected atoms.
The second component, that constitutes another novel contribution of this paper, implements an \emph{edge decoding} procedure, which predicts the presence of an edge between each pair of atoms and the corresponding  edge type.

\subsubsection{Atom Decoding}
The \emph{atom decoding} procedure receives in input a set of vectors $Z=\{\bz_v\}_{v \in [1,m]}$, where each $\bz_v$ represents an atom.
At the beginning of the generation phase, the $\bz_v$'s are sampled from the normal distribution \(\mathcal{N}\left(0, \bI\right)\), where $\bI$ is the identity matrix, while in training they are obtained from the distribution \( \mathcal{N}\left(\bmu_v,\bsigma^2_v\right)\) using the \emph{reparameterization trick}.
The \emph{atom decoding} returns for each $\bz_v$ a feature vector $\bbr_v \in \mathbb{R}^{s_n}$ jointly with its atom type $\tau_v$, where $s_n$ is the dimension of the features.
For atom decoding, we adopted the procedure provided in~\cite{rigoni2020conditional},  where the atom type assignment process is conditioned by the previously assigned atom types.
However, contrary to~\cite{rigoni2020conditional}, in the generation process we fix the $\alpha_0$ histogram for each iteration and we do not sample a new compatible histogram, because the sampling procedure negatively affects some metrics.
For more details on the decoding of the atoms, we refer the reader to Appendix D.

\subsubsection{Edge Decoding}
\begin{figure*}[t]
\centering
\includegraphics[width=1\textwidth]{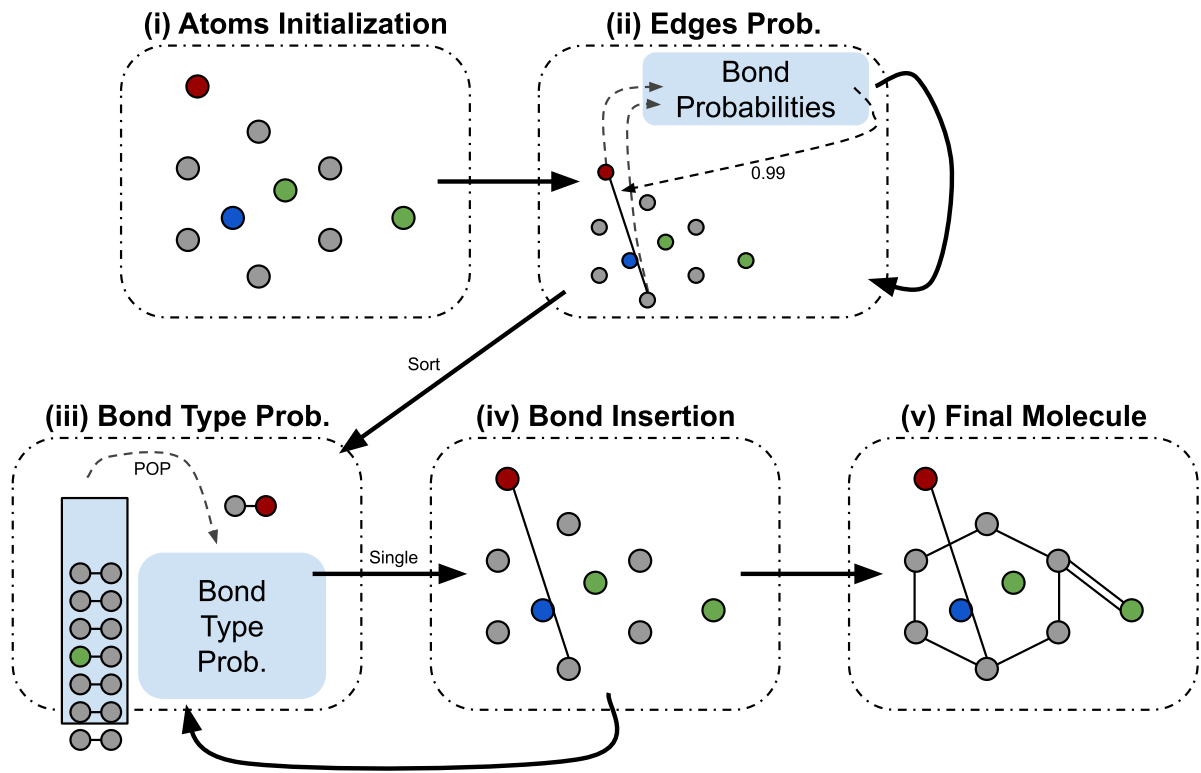}
\caption{Edge decoder. (\textit{i}) The atom representations are fed to (\textit{ii}) a neural network that computes the probability for each pair of atoms to be connected with an edge. 
Then, (\textit{iii}) the edges are sorted in descending probability order and are (\textit{iv}) iteratively added to the partial molecule according to their sampled edge type.
(\textit{v}) The process finishes when all the bonds are added to the  molecule.}
\label{fig:edges_decoder}
\end{figure*}

The \emph{edge decoding} procedure predicts the presence and type of an edge between each pair of atoms.
This procedure receives in input, for each node $v$, the feature vector $\bbr_v$ with its atom type $\tau_v$ returned by the \emph{atom decoding} step.
The process is illustrated in Fig.~\ref{fig:edges_decoder}.
Given a pair of atoms $(u, v)$ extracted from the set of initially unconnected atoms and their representations $\bbr_u$ and $\bbr_v$ (Fig.~\ref{fig:edges_decoder}-(i)), two neural networks are used to predict the probability that the two atoms are connected with a bond and the bond type, respectively (Fig.~\ref{fig:edges_decoder}-(ii)).
Specifically, using the compact notation $\leftrightarrow$ and $\stackrel{l}{\leftrightarrow}$ in order to represent  the existence of an edge and the existence of an edge of type $l$, respectively, the probability distribution $P\left(v \stackrel{l}{\leftrightarrow} u \mid \bphi_{v, u}\right)$ is:
\begin{equation}
    \underbrace{P\left(v \stackrel{l}{\leftrightarrow} u \mid \bphi_{v, u}, v \leftrightarrow u\right)}_\text{Edge type probability}  
\underbrace{P\left(v \leftrightarrow u \mid \bphi_{v, u}\right)}_\text{Edge probability},
\end{equation}
where
\begin{align}
\bphi_{v, u} &= \text{Concat}\left(\bs_v + \bs_u, \bs_v \odot \bs_u, \sum_{i=0}^{m-1} \bs_i\right) , \\
\bs_v &= \text{Concat}\left(\bbr_v, W(\tau_v) \right) \\
\bs_u &= \text{Concat}\left(\bbr_u, W(\tau_u) \right)
\end{align}
with $\odot$ the component-wise multiplication and $\text{Concat}(\cdot,\cdot)$ is the concatenation function. 

Note that $\sum_{i=0}^{m-1} \bs_i$ is a global order-invariant representation of all the atoms involved in the sum, and that the obtained matrix containing all the final probabilities is symmetric. In fact, this property is obtained thanks to $\bphi_{v, u}$ which is defined by only using order invariant representations of the couple of atoms $v$ and $u$.
We define the above probability functions as:
\begin{align}
&\begin{aligned}
    P&\left(v \leftrightarrow u \mid \bphi_{v, u}\right) = \\
    &=\frac{\bM^e_{v, u} \odot \exp \left[C\left(\bphi_{v, u}\right)\right]}
    {\bM^e_{v, u} \odot \exp \left[C\left(\bphi_{v, u}\right)\right] + 1}, 
\end{aligned}\\
&\begin{aligned}
    P&\left(v \stackrel{l}{\leftrightarrow} u \mid \bphi_{v, u}, v \leftrightarrow u\right) =\\
    &=\frac{\bM^t_{v, l, u} \odot \exp \left[L_{l}\left(\bphi_{v, u}\right)\right]}
    {\sum_{k} \bM^t_{v, k, u} \odot \exp \left[L_{k}\left(\bphi_{v, u}\right)\right]},
\end{aligned}
\end{align}
where both \func{C}{\phi} and \func{L_l}{\phi} are MLP neural networks and both $\bM^e_{v, u} \in \{0, 1\}^{m \times m}$ and $\bM^t_{v, k ,u} \in \{0, 1\}^{m \times d_e \times m}$ are binary masks.
Specifically, $\bM^e_{v, u}$ is a mask not allowing for self loops and edge duplication, while $\bM^t_{v, l, u}$ is a mask designated to remove the edges that do not respect the valences law.
Let $\nu_v$ and $\nu_u$ be the valences of atoms $v$ and $u$, respectively, and let $we(l)$ with $l \in \Theta_e$ be the function computing the edge weights as follows: 1 for \emph{single bonds}, 2 for \emph{double bonds}, and 3 for \emph{triple bonds}. Then $\bM^t_{v, l, u}$ is defined as:
\begin{equation}
\begin{aligned}
    \bM^t_{v, l, u} =1 \iff &\bM^e_{v, u}= 1 \,   \land\,    \nu_u \geq we(l)  \\
    & \land \,  \nu_v  \geq we(l). 
\end{aligned}
\end{equation}

At this point, all the pairs of atoms $(u, v)$, where their probabilities satisfy $P\left(v \leftrightarrow u \mid \bphi_{v, u}\right) > 0.5$, are ordered from the highest to the lowest probability value.
Then, the edges are iteratively added one at a time to the atoms according to the type of edge sampled from the distribution $P\left(v \stackrel{l}{\leftrightarrow} u \mid \bphi_{v, u}, v \leftrightarrow u\right)$.

If an edge that needs to be added does not respect the chemical constraints, then the model samples, without replacement, a new bond. An edge is discarded only when all the bond types violate the chemical constraints.

At the end of the process (Fig.~\ref{fig:edges_decoder}-(v)), all the bonds necessary for the validity of the molecule are completed by adding hydrogen atoms connected to all the atoms whose valences are not correct \footnote{We use a public online software (RDKit) for the construction of the molecules.}.

Note that our encoder and edges decoder  exhibit the invariant permutation property, while the atom decoder follows the canonical SMILES node order in order to 
fully exploit the conditioning on valence histograms, which allows to improve the performances on the considered metrics, as demonstrated in the ablation study described in Section~\ref{sec:ablation}.
It should be stressed that this dependence on the canonical SMILES node order does not affect the capability of the model since the task of generating new molecules
does not incur in the graph isomorphism problem. In fact, during the evaluation of the generated molecules, we compare canonical SMILES.

\subsection{Property Optimizer}
\label{sec:optimizer}
VAE models can incorporate an optimization component to tackle the \emph{structure optimization task}, which aims to drive the exploration of the latent space towards molecules that exhibit better properties.
In the search for new drugs this is very useful as it allows, given a molecule in input, to optimize its structure towards the desired property.
Given the set of $m$ sampled latent points $Z=\{\bz_v\}_{v \in [1,m]}$ and the histogram $\balpha_0$, the optimizer $O_p(Z,\balpha_0)$ predicts the property $p  \in \Theta_p$ of the molecule. 
Specifically, the input points $Z$ and the histogram $\balpha_0$ are used as input in the first part of the decoder process to generate, for each node $v$,
the feature vector $\bbr_v$ with its atom type $\tau_v$.
Then the optimizer predicts the property value as:
\begin{equation}
\begin{aligned}
    O_p(Z,\balpha_0)   = & \sigma\Biggl( \sum_{v=0}^{m-1}  \sigma\left(Q^1_p(\bx_v) \right)  \\
    &\odot \text{tanh}\left(Q^2_p(\bx_v) \right)\Biggr),
\end{aligned}
\end{equation}
where $\bx_v = F \left(\text{Concat}(\bbr_v, \tau_v) \right)$, 
$F$ is a single layer neural network with \emph{Leaky-ReLU} activation function, $Q^1_p$ and $Q^2_p$ are linear neural networks with a single layer specific for the property $p$, $\sigma()$ is the \emph{sigmoid} function and tanh$()$ is the \emph{hyperbolic tangent} function.
At test time, we sample an initial set of latent points $Z$ and an histogram $\balpha_0$ according to the training set distribution, then we use gradient ascent to reach a set $Z'$ of local optimal points.

\section{Training}
\label{sec:training}
Let us define the graph reconstructed by the decoder as $\tilde{G} = (\tilde{V}, \tilde{E}, \tilde{\Phi_e}, \tilde{\Phi_n})$, which can be represented by the predicted adjacency matrix $\tilde{\bA} \in \mathbb{R}_{[0,1]}^{m\times m}$, edge type tensor $\tilde{\bE} \in \mathbb{R}_{[0,1]}^{m \times m \times d_e}$, node attribute matrix $\tilde{\bF} \in \mathbb{R}_{[0,1]}^{m \times d_n}$ and vector of molecule properties $\tilde{\bp} \in \mathbb{R}^{d_p}$.
During generation, these matrices are obtained sampling from the decoder's output probabilities, while during training they are obtained using the \textit{argmax} function.
The order in which the vertices appear in the matrices is given by the ordering imposed by the \emph{SMILES}~\cite{weininger1988smiles} coding.\\
Let $\bmu$ and $\bsigma^2$ be the matrices formed respectively by the $m$ means $\{\bmu_v\}_{v \in [1,m]}$ and variances $\{\bsigma^2_v\}_{v \in [1,m]}$ predicted by the encoder.
Let $\bZ \sim \mathcal{N}\left(\bmu,\bsigma^2 \right)$ be the matrix formed by the $m$ latent space sampled points $Z=\{\bz_v\}_{v \in [1,m]}$, then the loss function to minimize is  $\mathcal{L}(G, \tilde{G},\bmu, \bsigma^2)$, defined as:
\begin{equation}
\begin{aligned}
   \mathcal{L}(G, \tilde{G},\bmu, \bsigma^2) = &\mathcal{L}_{rec}(G, \tilde{G}) + \lambda_1 \mathcal{L}_{lt}(\bmu, \bsigma^2) \\
   &+\lambda_2 \mathcal{L}_{opt}(\bp, \tilde{\bp}),
\end{aligned}
\end{equation}
where \(\lambda_1\) and \(\lambda_2\) are trade-off hyper-parameters,
\(\mathcal{L}_{rec}\) is the reconstruction loss, \(\mathcal{L}_{lt}\) is the variational autoencoder \emph{Kullback–Leibler} (KL) loss, and \(\mathcal{L}_{opt}\) is the loss defined on the basis of the property to predict, referred to as \emph{optimization loss} in the literature since it can be used to drive the generation of molecules towards molecules with better values for the property of choice.
\noindent In more detail, $\mathcal{L}_{rec}$ is defined as: 
\begin{equation}
\begin{aligned}
    \mathcal{L}_{rec}(G, \tilde{G}) =& \mathcal{L}_{a}(\bF, \tilde{\bF}) + \mathcal{L}_{b}(\bA, \tilde{\bA}) \\
    &+ \mathcal{L}_{tb}(\bE, \tilde{\bE}), 
\end{aligned}
\end{equation}
where $\mathcal{L}_{a}$, $\mathcal{L}_{tb}$, and $\mathcal{L}_{b}$ are cross-entropy losses:
\begin{align}
&\begin{aligned}
    \mathcal{L}_{a}(\bF, \tilde{\bF}) = -\sum_{v=0}^{m-1} \sum_{t=0}^{d_n-1}  \log(\tilde{\bF}_{v,t}) \odot \bF_{v,t},
\end{aligned} \\
&\begin{aligned}
    \mathcal{L}_{tb}(\bE, \tilde{\bE}) = & -\sum_{v=0}^{m-1} \sum_{u=0}^{m-1} \underbrace{\bA_{v,u}}_\text{Teacher forcing} \\
    &  \odot  \Biggl(\sum_{t=0}^{d_e-1} \log({\tilde{\bE}_{v,u,t}}) \odot \bE_{v,u,t} \Biggr)
\end{aligned} \\
&\begin{aligned}
    \mathcal{L}_{b}(\bA, \tilde{\bA}) = & -\sum_{v=0}^{m-1} \sum_{u=0}^{m-1} \Biggl(\log(1-\tilde{\bA}_{v,u}) \\
    & \odot (1-\bA_{v,u}) +  \log(\tilde{\bA}_{v,u}) \Biggr)
\end{aligned}
\end{align} 
Note that the loss $\mathcal{L}_{tb}$ uses \emph{teacher forcing} to calculate the bond-type loss only where the bond exists in the molecule.
The variational autoencoder loss \emph{Kullback–Leibler} is defined as:
\begin{equation}
\begin{aligned}
    \mathcal{L}_{lt}(\bmu, \bsigma^2) =&
    - \frac{1}{2} \sum_{i=0}^{m-1} \sum_{j=0}^{s_{\text{lt}}-1} 1 + \log(\bsigma^2_{i,j}) \\
    &- \bmu_{i,j}^2 - \bsigma^2_{i,j} ,
\end{aligned}
\end{equation}
while the optimization loss $\mathcal{L}_{opt}$ is defined as: 
\begin{align}
    \mathcal{L}_{opt}(\bp, \tilde{\bp}) = \sum_{p=1}^{d_p}  \dfrac{(\tilde{\bp_p} - \bp_p)^2}{2}.
\end{align}

\section{Related Works}
\label{sec:works}
Follow related works that tackle the \emph{distribution learning task}.

Focusing on VAE approaches, Character VAE~\cite{gomez2018automatic} exploits, in input and output, SMILES strings describing the structure of the molecule.  The encoder uses a convolutional network and the decoder uses a gated recurrent unit (GRU). 
Grammar VAE~\cite{DBLP:conf/icml/KusnerPH17} adds to Character VAE a \emph{context-free grammar}, to guide the correct generation of SMILES strings.
Syntax Directed VAE~\cite{DBLP:conf/iclr/DaiTDSS18} improves Grammar VAE using a more expressive grammar,the \emph{attribute grammar}~\cite{knuth1968semantics}, which aims to generate strings that not only are syntactically valid, but also semantically reasonable.
Junction Tree VAE~\cite{DBLP:conf/icml/JinBJ18} represents molecules using graphs, composed of chemical substructures that are extracted from the training set. New molecular graphs are obtained by first generating a tree-structured scaffold formed by substructures, and then combining the substructures together using a graph message passing network.
HierVAE~\cite{jin2020hierarchical}  uses a hierarchical graph encoder-decoder that employs large and flexible graph motifs as basic building blocks to encode and decode a molecule.
\cite{podda2020deep} generates molecules fragment by fragment instead of atom by atom, using a language model based on the SMILES representation of molecules.
Regularized Graph VAE~\cite{DBLP:conf/nips/MaCX18} casts the molecule generation problem as a constrained optimization problem, where chemical constraints are encoded in the VAE loss function.
Constrained Graph VAE (CGVAE)~\cite{DBLP:conf/nips/LiuABG18} encodes in the latent space single atoms rather than whole molecules. To generate a molecule, first the model samples several nodes in the latent space and assigns them an atom type using a linear classifier; then it connects them using a constrained breadth first algorithm. Both the encoder and decoder are implemented by a gated graph sequence neural network, that makes the model computationally heavy to train.
CCGVAE~\cite{rigoni2020conditional} extends CGVAE conditioning only the decoder with the histogram of valences of the molecule's atoms. 
In particular, the decoder takes in input several nodes sampled in the latent space together with a histogram of valence, and for each of those points, it assigns its atom type applying an iterative procedure that depends on the histogram of valences.

CCGVAE model is the one most related to our proposal as we also condition the decoder with the histogram of valence information.
However, our RGCVAE architecture is completely different from that of CCGVAE.
In fact, in our model, we have an encoder that exploits the newly defined representations of molecules using the new relational GIN component.
Furthermore, the decoder does not contain a convolutional operator which is instead present in CCGVAE.
Finally, our optimization component does not optimize points directly from latent space, but predicts scores from the internal states of the decoder component.

Focusing on the generative adversarial approach, MolGAN~\cite{de2018molgan}, learns via reinforcement learning to directly reconstruct small organic molecules, by predicting directly the atoms type, and the existence of bonds (and their types).
NeVAE~\cite{DBLP:conf/aaai/SamantaDJCGR19} predicts the spatial coordinates of the atoms of the generated molecules, and optimizes them to improve its structures. 

Focusing on the Flow-based models,  GraphAF~\cite{shi2020graphaf} combines the advantages of both autoregressive and flow-based approaches in order to model the data density estimation, while MolGrow~\cite{DBLP:conf/aaai/KuznetsovP21} uses a hierarchical normalizing flow-based model for generating molecular graphs.
MoFlow~\cite{DBLP:conf/kdd/ZangW20} generates molecular graphs by first generating bonds through a Glow based model, then by generating atoms given bonds by a novel graph conditional flow-based model.

Regarding the evaluation process adopted in this work, we embrace the same testbed environment defined in~\cite{rigoni2020systematic} to compare our results with many other state-of-the-art models objectively.

\section{Experiments}
\label{sec:experiments}
Following~\cite{rigoni2020systematic, rigoni2020conditional}, we compared our RGCVAE to several state-of-the-art Variational Autoencoder proposals considering different metrics on two widely adopted datasets.
Specifically, focusing on the \emph{distribution learning task}, for each model we assessed the ability to reconstruct the input molecules and the ability to generate new ones.

\subsection{Experimental setup}
Following~\cite{rigoni2020systematic}, we evaluated our model on two widely used datasets: QM9~\cite{ruddigkeit2012enumeration,ramakrishnan2014quantum}, which includes $\sim$134,000 organic molecules with at most 9 atoms, and  ZINC~\cite{irwin2005zinc},  which includes 250,000 drug-like molecules with at most 38 atoms.
More details on the datasets are reported in Appendix A.
We use for each model the same training, validation and test splits as in~\cite{rigoni2020systematic}.

We considered the following metrics:
\begin{enumerate*}[label=(\roman*)]
    \item \emph{Reconstruction} that, given an input molecule and a set of generated molecules, computes the percentage of generated molecules that are equal to the one in input. This is calculated on the \emph{test set};
    \item \emph{Validity} that, given a set of generated molecules, represents the percentage of them that is valid, i.e. that represent actual molecules;
    \item \emph{Novelty} that represents the percentage of  generated molecules not in the training set;
    \item \emph{Uniqueness} that represents (in percentage) the ability of the model to generate different molecules in output, and is computed as the size of the unique set of valid generated molecules divided by the total number of valid generated molecules;
    \item \emph{Diversity} that measures how much the generated molecules are different from those in the training set (comparing randomly selected substructures).

    \item \emph{Quantitative Estimation Drug-likeness (QED)} which indicates in percentage how likely it is that the molecule is a good candidate to become a drug. We used this metric to evaluate the optimization procedure.
\end{enumerate*}
\emph{Reconstruction} is computed on $5,000$ test set molecules encoded and decoded 20 times, while the other metrics are computed sampling $20,000$ points from the standard normal distribution and decoding each point once. 

Notice that in our model we deal with molecules defined with \emph{stereochemistry} information, and for this reason, during the evaluation of the \emph{Reconstruction} metric, we always compare the \emph{canonical} SMILES representation. Some works in literature compare graph representations that do not contain such information or they clean the SMILES representation before training/evaluation phase. We refer the reader to Section \ref{sec:ablation} to more details.

The experiments on the QM9 dataset were performed on a PC with an Intel(R) i9 9900k CPU, 32 GB of RAM and a Quadro RTX 4000 GPU with 8GB of memory. For the experiments on the ZINC dataset we used a server with 2 Intel(R) Xeon(R) E5-2650a CPUs, 160 GB of RAM and a Tesla T4 GPU with 15GB of memory.
Regarding the model selection process, we  started from a model where each neural network component of RGCVAE was implemented as a linear transformation and then, according to the obtained reconstruction error on the validation set, we iteratively added hidden layers to each neural network. 
When adding depth to the networks did not improve the reconstruction metric anymore, we used a grid search to find the best value for the $\lambda_1$ parameter which controls the importance of the \emph{Kullback–Leibler} divergence in the model loss.
The network, as shown in Section \ref{sec:ablation}, is sensitive with respect to this parameter that trades off between reconstruction on one hand, and uniqueness and novelty on the other hand.

Due to limitations in the available computational resources, we did not fine-tune the other parameters. 
The results presented in this section are computed on the test set with the best model obtained in the model selection process.
More details on the model selection procedure are reported in Appendix E.

The approaches proposed in literature are evaluated using different metrics and on different datasets or variations of them.

For this reason, it was necessary to reproduce the baseline results to calculate all the metrics in the datasets considered in this paper. 
For each baseline model, we used the official code and set the hyper-parameters to the values specified by the authors. 
Consequentially, we have trained and tested each baseline only on the datasets originally considered by the authors and adopted in this work.
In Appendix B, we report for a reference the results obtained by the baseline models in the datasets considered in this paper that where originally not considered by the authors, while Appendix C presents results considering the molecules properties.

\begin{table*}
\begin{center}
\begin{minipage}{1\linewidth}
\caption{Average and standard deviation (where applicable) of different evaluation metrics on the QM9 datasets.}
\label{tab:results_qm9}
\begin{tabular}{ lcccccc }
\toprule
\textbf{Model}      &\textbf{Reconstruction}    &\textbf{Validity}&\textbf{Novelty}&\textbf{Uniqueness}   &\textbf{Diversity}  &\textbf{Time}    \\
\midrule
\mr{Graph VAE}                          &13.6          &80.1          &45.6             &\mr{88.1}         &66.2              &\mr{01m}      \\
                                        &\(\pm\)34.3   &\(\pm\)32.4   &\(\pm\)49.8      &                   &\(\pm\)28.0       &             \\
\mr{Reg. GVAE}                          &7.3           &91.8          &49.8             &\mr{77.1}             &68.7            &\mr{01m}    \\
                                        &\(\pm\)26.0   &\(\pm\)27.5   &\(\pm\)50.0      &                   &\(\pm\)25.6       &             \\
\mr{CGVAE} &24.5                        &\dcb{}100       &92.8          &\mr{98.3}         &76.1              &\mr{21m}       \\
                                        &\(\pm\)27.9   &\(\pm\)0        &\(\pm\)19.1   &                   &\(\pm\)22.6       &              \\
\mr{CCGVAE}&55.4                        &\dcb{}100       &88.5          &\mr{93.2}         &79.2              &\mr{1.5h}         \\
                                        &\(\pm\)49.7   &\(\pm\)0        &\(\pm\)31.9   &                   &\(\pm\)22.0       &              \\                             
\mr{\textbf{RGCVAE (ours)}}             &94.8           &99.9          &96.4             &\mr{94}           &87.7              &\mr{01m}        \\            
                                        &\(\pm\)22.2    &\(\pm\)3.5       &\(\pm\)18.7    &                 &\(\pm\)19.2       &           \\
\bottomrule
\end{tabular}
\end{minipage}
\end{center}
\end{table*}

\begin{table*}
\begin{center}
\begin{minipage}{1\linewidth}
\caption{Average and standard deviation (where applicable) of different evaluation metrics on the ZINC dataset.}
\label{tab:results_zinc}
\begin{tabular}{ lcccccc }
\toprule
\textbf{Model}      &\textbf{Reconstruction}    &\textbf{Validity}&\textbf{Novelty}&\textbf{Uniqueness}   &\textbf{Diversity}  &\textbf{Time}    \\
\midrule
\mr{Char. VAE}                      &25.3              &0.9             &\dcb{}100      &\mr{91.4}          &98.2               &\mr{07m}    \\  
                                    &\(\pm\)43.5       &\(\pm\)9.6      &\(\pm\)0       &                   &\(\pm\)7.0         &             \\ 
\mr{Gram. VAE}                      &55.8              &5.1             &\dcb{}100      &\mr{94.6}          &\dcb{}99.2          &\mr{21m}         \\
                                    &\(\pm\)49.7       &\(\pm\)23.0     &\(\pm\)0       &                   &\(\pm\)4.5         &                \\
\mr{SD VAE}                         &\dcb{}77.4        &19.0            &\dcb{}100      &\dcb{}             &93.6               &\mr{24m}        \\
                                    &\(\pm\)41.8       &\(\pm\)39.2     &\(\pm\)0       &\dcb{}\mrm{100}    &\(\pm\)18.5        &                \\
\mr{JT VAE}                         &50.2              &99.6          &99.9             &\mr{99.7}          &33.0                &\mr{55m}         \\
                                    &\(\pm\)50.0       &\(\pm\)6.35    &\(\pm\)1.2      &                   &\(\pm\)21.78       &               \\
\mr{CGVAE}   &0.4                   &\dcb{}100          &\dcb{}100      &\mr{99.9}      &66.0               &\mr{15.5h}                   \\
                                    &\(\pm\)5.9        &\(\pm\)0       &\(\pm\)0        &                   &\(\pm\)22.8       &               \\
\mr{CCGVAE} &22.2                   &\dcb{}100          &\dcb{}100     &\mr{92.8}       &80.0               &\mr{21.5h}                         \\
                                    &\(\pm\)41.5        &\(\pm\)0       &\(\pm\)0       &                   &\(\pm\)16.7        &               \\
\mr{\textbf{RGCVAE (ours)}}         &79.7               &99.9            &100           &\mr{61.1}          &98.9               &\mr{12m}                 \\            
                                    &\(\pm\)40.2        &\(\pm\)1.2     &\(\pm\)0       &                   &\(\pm\)5.0         &                    \\
\bottomrule
\end{tabular}
\end{minipage}
\end{center}
\end{table*}

\subsection{Reconstruction and Generation}
Table~\ref{tab:results_qm9} and Table~\ref{tab:results_zinc} report the average and standard deviation of the results obtained by the models on the QM9 and ZINC datasets, respectively~\footnote{Notice that distributions are skewed, which implies that it is possible to obtain high average values with large standard deviations.}. 
Moreover, they report the computational time required to perform one training epoch.
In general, all the metrics adopted in this work, capture different but important aspects of the molecule generation process. Ideally, a model should perform well in all such metrics. 
Having even only one of them very low indicates a strong limitation of such model. 
In particular, VAE models usually trade-off the capability of the model of reconstructing the molecule in input with the capability of the model of generating new valid, novel and useful molecules. 
For instance, MOSES~\cite{polykovskiy2018molecular}, that does not adopt the reconstruction metric in its evaluation process, presents very high validity, novelty, and uniqueness values for Character VAE~\cite{gomez2018automatic} and AAE~\cite{kadurin2017cornucopia}. However, our evaluation of reconstruction\footnote{We implemented the reconstruction function and evaluated it on the MOSES official code.} for the models released by MOSES' authors returned values very close to zero for each of these two models.
In fact, authors  report in the official MOSES's GitHub repository that the obtained models suffer by the posterior collapse problem. This means that the encoding process is not contributing to the generation process, which raises many doubts on how much the generated molecules follow the training set distribution.

In our evaluation process, for each model, we have used the official code and the hyper-parameter values provided by the authors. However, we could not reproduce the 76.7\% reconstruction of the JTVAE model. Many users pointed out bugs in the code, and they also claimed not being able to achieve the values reported in the original work. We thus consider the results in the JTVAE paper not reproducible.

Considering our results, we notice that methods that are based on SMILES strings instead of graphs (i.e. Character VAE, Grammar VAE and Syntax Directed VAE) perform poorly. Specifically, the validity of the generated strings is very low, i.e. the networks struggle to generate strings that are valid SMILES. Thus, even though on some metrics these models perform pretty well, they are not well suited for the task of molecule generation.
Methods that use the graph representation of molecules show higher performance. Among them, our model (RGCVAE) presents very good results in both the QM9 and ZINC datasets.
In particular, on the QM9 dataset, when compared to the best baseline model  (CCGVAE considering all the metrics), RGCVAE improves its \emph{Reconstruction} by almost 39\%.
On the ZINC dataset, RGCVAE reconstruction is 29.5\% higher than JTVAE (3\% considering the original results reported by the authors) and 55.5\% higher than CCGVAE.
Notwithstanding the good \emph{Reconstruction} results, RGCVAE also shows high values in the
 \emph{Validity}, \emph{Novelty}, \emph{Uniqueness} and \emph{Diversity} metrics  on both the datasets.
The only metric where RGCVAE is not among the best performing methods is the \emph{Uniqueness} on the ZINC dataset, which is lower compared to JTVAE and CCGVAE. 
We argue that this relatively low \emph{Uniqueness} value is a consequence of the well-formed latent space of the model amenable to the high \emph{Reconstruction} value. In Section~\ref{sec:ablation}, we show that sacrificing the model \emph{Reconstruction} capabilities the model reaches higher \emph{Uniqueness} values.
Notice, however, that both CCGVAE and JTVAE show a much lower \emph{Diversity} compared to RGCVAE, which indicates that the generated molecules are built with really (98.9\% on ZINC) different sub-structures than those that form the molecule's training set. 

Concerning the computational efficiency, we can observe that the models that use the SMILES representation, i.e. the first three models in each table, tend to be faster to train than the ones that use the molecular graph representation, i.e. all the others. However, as stated before, their predictive performances are generally low.
The high computational time and resources required by some models for completing a training epoch, e.g. CGVAE and CCGVAE, limits the number of epochs that can be performed for training them in a reasonable amount of time. This may hurt their predictive performances. 
While Graph VAE, Regularized GVAE 

tend to be fast, Graph VAE and Regularized GVAE present low \emph{Reconstruction} and \emph{Novelty} values.

From these tables, we can notice that our proposed RGCVAE model is \textbf{among the fastest} methods.
Overall, our proposed model is very efficient while being at the-state-of-the-art in several metrics.

\subsection{Property Optimization}
This task aims to optimizing molecules structures towards desired properties.
Following~\cite{DBLP:conf/icml/JinBJ18}, starting from a set of points sampled in the latent space and an initial histogram of valences, the optimization algorithm performs gradient ascent moving the representations toward molecules that maximizes the \emph{QED} property, as described in Section~\ref{sec:optimizer}.

\begin{figure*}[t]
\centering
\begin{picture}(\textwidth,100)
\put(0,0){\includegraphics[scale=0.4]{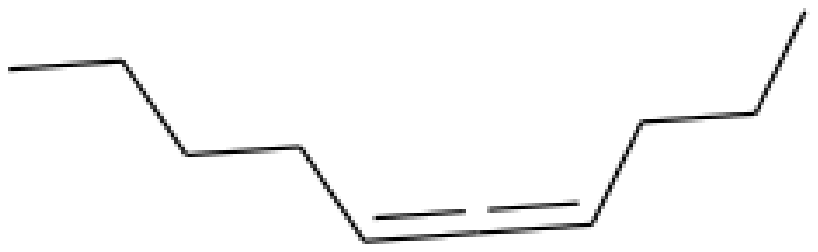}}
\put(\textwidth/4,0){\includegraphics[scale=0.4]{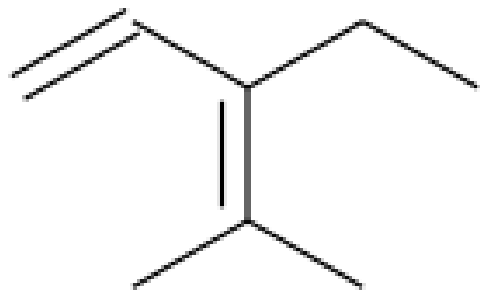}}
\put(\textwidth/4*2,0){\includegraphics[scale=0.4]{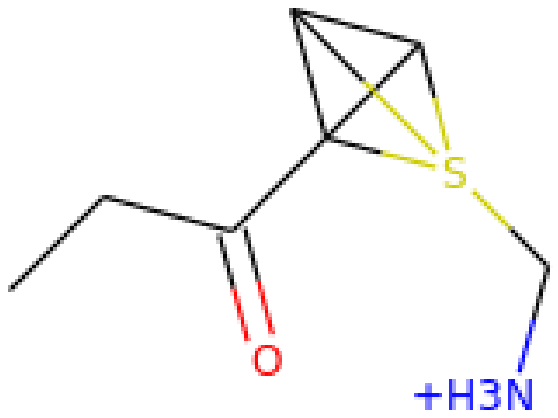}}
\put(\textwidth/4*3,0){\includegraphics[scale=0.4]{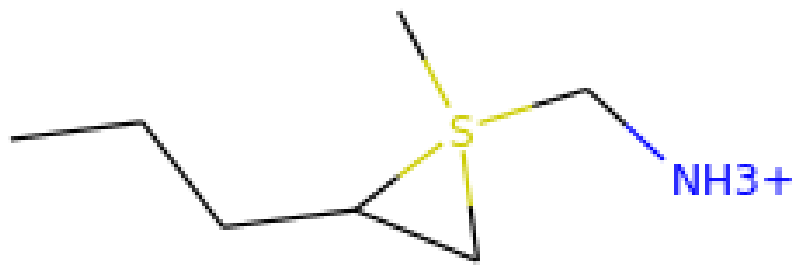}}
\put(45,10){(i) 0.447}
\put(\textwidth/4+45pt,10){(ii) 0.497}
\put(\textwidth/4*2+45pt,10){(iii) 0.526}
\put(\textwidth/4*3+45pt,10){(iv) 0.583}
\end{picture}
\caption{Example of QED-directed optimization performed in the model latent space trained on the ZINC dataset. (i) is the initially sampled molecule, while (ii)-(iv) are the  molecules obtained at each optimization step.}
\label{fig:optimization}
\end{figure*}

In Fig.~\ref{fig:optimization}, we present an example of \emph{QED} optimization on the ZINC dataset.
Starting from left, we can see the sampled molecule and the molecules obtained at each optimization time step. 
We report the real \emph{QED} values, from which we can see that the model is able to improve the molecule structure towards those that improve the \emph{QED} value.

\section{Ablation Study}
\label{sec:ablation}
In this section, we present an ablation study to validate the different components of RGCVAE.
Specifically, we analyze the impact on the predictive performance of the atom encoding (different choices are possible), of the Relational GIN over a standard GIN and the R-GCN~\cite{schlichtkrull2018modeling} in the encoder (Section~\ref{sec:encoder}), and of the use of the histograms of valence (Section~\ref{sec:decoder}).
\begin{table*}
\begin{center}
\begin{minipage}{0.85\linewidth}
    \caption{Minimum reconstruction errors that is possible to achieve with different graph representations.\label{tab:ablation_bounds}}
    \begin{tabular}{ ccccccc }
    \toprule
    \textbf{Representation} &\textbf{Atom}&\textbf{Valence}&\textbf{Charge}&\textbf{Chiral}  &\textbf{\%QM9}    &\textbf{\%ZINC}     \\ 
    \midrule
    1           &\CheckmarkBold   &\XSolidBold      &\XSolidBold    &\XSolidBold   &22.83           &72.32                      \\
    2           &\CheckmarkBold   &\CheckmarkBold   &\CheckmarkBold   &\XSolidBold   &1.70            &63.08            \\
    3           &\CheckmarkBold   &\CheckmarkBold   &\CheckmarkBold   &\CheckmarkBold   &1.70            &7.76                        \\
    \bottomrule
    \end{tabular}
\end{minipage}
\end{center}
\end{table*}

In our work, we deal with molecules that include the \emph{stereochemistry} relative spatial arrangement of atoms in the molecules.
For this reason, when the molecule is converted as a graph representation, it is necessary to store, learn, and predict these information to build back the original molecule, otherwise sometimes it is not possible to reconstruct the original molecule.
This makes the learning process more difficult.
Notice that many models in literature {\bf do not} consider this information, and they just compare molecules using their graph representations or SMILES strings~\cite{DBLP:conf/kdd/ZangW20}, where the \emph{stereochemistry} information is removed, leading to higher \emph{Reconstruction} values than those reported in this work.
Moreover, this information is often only reported in the code, not in the paper.
In order to reach the maximum \emph{Reconstruction} value and do not occur in low values due to the lack of information in the molecule's graph representation, we have considered three atom representations:
\begin{enumerate*}
    \item atom type, e.g. ``\textbf{C}'' for  carbon;
    \item atom type, total valence and formal charge, e.g. ``\textbf{C4(0)}'' for carbon atom with total valence 4 and formal charge 0;
    \item atom type, total valence, formal charge, presence of chiral property, e.g. ``\textbf{O3(1)0}'' for an oxygen atom with valence 3, formal charge 1 and without the chiral property. 
\end{enumerate*}   

Table~\ref{tab:ablation_bounds} reports the minimum \emph{Reconstruction} error that is possible to achieve using these representations on the two adopted datasets.
It is clear that the third representation is the one that better captures the molecule structure information (including the \emph{stereochemistry} information).
In fact, the third representation provided the best performance for our model.
Moreover, we verified the importance of using the Relational GIN in the encoder, and the histograms in the decoding procedure.
\begin{sidewaystable*}
\begin{center}
\begin{minipage}{0.9\linewidth}
    \caption{This table reports the ablation study results obtained by our RGCVAE model.}
    \label{tab:ablation}
    \begin{tabular}{ccccccccccccc}
    \toprule
    \multicolumn{3}{c}{}&\multicolumn{5}{c}{\textbf{QM9}} & \multicolumn{5}{c}{\textbf{ZINC}}  \\
    \cmidrule(lr{4pt}){4-8} \cmidrule(lr{4pt}){9-13}
    \textbf{Rep.}& \textbf{Hist.} & \textbf{Encoding Network}&\textbf{Rec.}&\textbf{Val.}&\textbf{Nov.}&\textbf{Uniq.}&\textbf{Div.}&\textbf{Rec.}&\textbf{Val.}&\textbf{Nov.}&\textbf{Uniq.}&\textbf{Div.} \\
    \midrule
    \mr{1}      &\mr{\XSolidBold}       &\mr{RGIN}   &68.8           &100            &77.9           &\mr{84.1}      &69.95           &20.11           &99.9           &100            &\mr{1.9}       &100           \\
                &                       &            &\(\pm\)46.3    &\(\pm\)0       &\(\pm\)41.5    &               &\(\pm\)27.6    &\(\pm\)40.1    &\(\pm\)1.7     &\(\pm\)0          &               &\(\pm\)0.1      \\
    \mr{1}      &\mr{\CheckmarkBold}    &\mr{RGIN}   &76.2           &99.9            &95.8           &\mr{90.4}      &87.7           &27.1           &97.0           &100            &\mr{90.3}      &96.5           \\
                &                       &            &\(\pm\)42.6    &\(\pm\)3.2     &\(\pm\)20.1    &               &\(\pm\)19.4    &\(\pm\)44.5    &\(\pm\)17.0     &\(\pm\)0       &               &\(\pm\)9.2    \\

    \mr{3}      &\mr{\XSolidBold}       &\mr{RGIN}   &85.9           &100            &78.1           &\mr{83.9}      &69.6           &47.4           &100            &100            &\mr{2.4}       &100           \\
                &                       &            &\(\pm\)34.8    &\(\pm\)0       &\(\pm\)41.4    &               &\(\pm\)28.2    &\(\pm\)49.9    &\(\pm\)0       &\(\pm\)0       &               &\(\pm\)0    \\

    \mr{3}      &\mr{\CheckmarkBold}    &\mr{RGIN}   &94.8           &99.9            &96.36           &\mr{94.0}      &87.7           &79.7           &99.9           &100            &\mr{61.1}      &98.9           \\
                &                       &            &\(\pm\)22.21   &\(\pm\)3.5     &\(\pm\)18.7    &               &\(\pm\)19.2    &\(\pm\)40.2      &\(\pm\)1.2     &\(\pm\)0       &               &\(\pm\)4.98   \\

    \mr{3}      &\mr{\CheckmarkBold}    &\mr{GIN}    &41.6           &99.3           &94.1           &\mr{87.2}      &87.47           &54.0           &100           &100            &\mr{42.2}         &99.8           \\
                &                       &            &\(\pm\)49.3    &\(\pm\)8.5     &\(\pm\)23.5     &               &\(\pm\)19.9    &\(\pm\)49.8    &\(\pm\)0     &\(\pm\)0       &               &\(\pm\)2.14    \\

    \mr{3}      &\mr{\CheckmarkBold}    &\mr{R-GCN}  &86.0           &99.8            &95.5           &\mr{89.1}      &88.5           &64.9           &100           &100            &\mr{81.53}      &96.44           \\
                &                       &            &\(\pm\)34.7    &\(\pm\)4.1       &\(\pm\)20.76    &            &\(\pm\)19.41    &\(\pm\)47.7    &\(\pm\)0     &\(\pm\)0       &               &\(\pm\)9.41   \\
    \bottomrule
    \end{tabular}
\end{minipage}
\end{center}
\end{sidewaystable*}

Table~\ref{tab:ablation} reports the results obtained by different versions of our model, including or not the histogram of valences, the use of a standard GIN network or a standard R-GCN network over the novel Relational GIN for the encoder, and using either representation 1 or 3, i.e. the simplest and the most sophisticated representation. 

We evaluated the effectiveness of the Relational GIN only on the best performing representation 3.
To make a fair comparison, we have selected the model that has the higher \emph{Reconstruction} value on the validation set among 200 epochs for each dataset.

From tables it is evident that using the histogram of valences to condition the decoder improves the model ability in both the reconstruction and the generation tasks.
In fact, on the QM9 dataset, it improves the \emph{Reconstruction} values by  $\sim$7\% when using representation 1, and above $8\%$ when using representation 3. A similar improvement is observed also for the \emph{Uniqueness} metric, although in a reduced form for representation 1.
We can observe the same behaviour on the ZINC dataset.

Notice that, using both representation 1 and 3, if the model does not use the histogram of valences, the \emph{Uniqueness} values are very low.
We argue that these improvements are due to the fact that using the histogram of valences: \emph{(i)} in reconstruction, drives the generation process toward molecules with a valence distribution closer to the ones of the molecules in the training set;
\emph{(ii)} in generation, drastically decreases the KL divergence loss, which means that the latent space is closer to the standard normal distribution, i.e.  the distribution from which the generated latent representations are sampled.

From the results, it is also clear that the Relational GIN dramatically improves the performance over the versions using the standard GIN and the R-GCN.
Regarding the expressiveness of the different types of atom representation, as it was expected, the results improve with  atom representation 3.

Finally, in Table \ref{tab:ablation_lambda} we report the results of our model when changing value of the $\lambda_1$ hyper-parameter. As expected, increasing the value leads to lower values of {\it Reconstruction} while improving in many cases the other metrics. 
\begin{table*}
\begin{center}
\begin{minipage}{0.85\linewidth}
    \caption{This table reports the ablation study results obtained by our RGCVAE model as the $\lambda_1$ hyper-parameter value changes.}
    \label{tab:ablation_lambda}
    \begin{tabular}{ ccccccccccc }
    \toprule
    &\multicolumn{5}{c}{\textbf{QM9}} & \multicolumn{5}{c}{\textbf{ZINC}}  \\
    \cmidrule(lr{4pt}){2-6} \cmidrule(lr{4pt}){7-11}
    $\bm{\lambda_1}$&\textbf{Rec.}&\textbf{Val.}&\textbf{Nov.}&\textbf{Uniq.}&\textbf{Div.}&\textbf{Rec.}&\textbf{Val.}&\textbf{Nov.}&\textbf{Uniq.}&\textbf{Div.} \\
    \midrule
    \mr{0.01}  &96.5           &99.5            &95.4           &\mr{90.9}      &79.9            &79.7           &99.9          &100            &\mr{61.1}       &98.9           \\
               &\(\pm\)18.3    &\(\pm\)7.9      &\(\pm\)20.9    &               &\(\pm\)23.2     &\(\pm\)40.2    &\(\pm\)1.2    &\(\pm\)0    &               &\(\pm\)5.0      \\
    \mr{0.05}  &94.8          &99.9            &96.4           &\mr{94.0}      &87.7            &62.2           &99.9           &100            &\mr{99.8}      &92.9           \\
               &\(\pm\)22.2   &\(\pm\)3.5      &\(\pm\)18.7   &              &\(\pm\)19.2       &\(\pm\)48.5    &\(\pm\)1.9     &\(\pm\)0       &               &\(\pm\)13.0    \\
    \mr{0.1}   &93.3           &99.9            &97.1           &\mr{95.2}      &90.4           &55.62           &99.9            &100            &\mr{86.5}       &97.9           \\
               &\(\pm\)25.0    &\(\pm\)3.0     &\(\pm\)16.7    &               &\(\pm\)17.4     &\(\pm\)49.7    &\(\pm\)0.7       &\(\pm\)0       &               &\(\pm\)7.0    \\
    \mr{0.2}   &83.5           &99.9            &96.6           &\mr{87.3}      &93.0           &40.0           &99.9           &100            &\mr{99.9}      &94.9           \\
               &\(\pm\)37.18    &\(\pm\)3.4     &\(\pm\)18.2    &               &\(\pm\)15.2    &\(\pm\)49.0    &\(\pm\)1.0     &\(\pm\)0       &               &\(\pm\)10.2   \\
    \bottomrule
    \end{tabular}
\end{minipage}
\end{center}
\end{table*}

\section{Conclusions and Future Works}
\label{sec:conclusions}
This paper introduced RGCVAE, a VAE deep generative model for molecule generation. 
The main features of RGCVAE are: (i) an improved reconstruction error with respect to state-of-the-art VAE models; (ii) a very efficient runtime, both in training and in generation.
These features are obtained thanks to the novel contributions of this paper, i.e. an extension of the GIN model for relational data, RGIN, exploited during the encoding phase, and a new and efficient decoder which allows to effectively generate novel molecules. 
In order to validate the usefulness of the different information sources exploited by the model, we have performed an ablation study that confirmed the goodness of the choices made. 
We have compared our model results against several VAE models in literature, considering many metrics on two widely adopted datasets.
Results show that RGCVAE perform state-of-the-art molecule generation performance while being significantly faster to train.

Future works will aim at improving the model by understanding how to embed further knowledge priors into both the encoder and decoder.

\clearpage
\bibliographystyle{sn-chicago}
\bibliography{ref.bib}

\begin{appendices}
    \section{}
    \subsection{Dataset Details}
    \begin{table*}
\small
\begin{center}
\begin{minipage}{0.75\linewidth}
    \caption{Statistics about the QM9 and ZINC datasets.}
    \label{tab:datasets}
    \begin{tabular}{ ccccc }
        \toprule
        \textbf{Dataset} & \textbf{\#Molecules}& \textbf{\#Atoms}& \textbf{\#Atom Types} & \textbf{\#Bond Types}\\
        \midrule
        QM9     & 134K      & 9         & 4     & 3     \\
        ZINC    & 250K      & 38        & 9     & 3     \\
        \bottomrule
    \end{tabular}
\end{minipage}
\end{center}
\end{table*}
    As presented in the main article, Table~\ref{tab:datasets} synthetically reports the main characteristics of the QM9 and ZINC datasets used for the evaluations. 
    Both use the same number of bond types, but they differ in the number of atom types, i.e. 4 for QM9 and 9 for ZINC, and atom per molecules, i.e. at maximum 9 for QM9 and 38 for ZINC.

    \section{}
    \subsection{Model Results}
    \begin{sidewaystable*}
\begin{center}
\begin{minipage}{0.9\linewidth}
\caption{Average and standard deviation (where applicable) of different evaluation metrics on the QM9 and ZINC data-sets. 
The symbols '\ddag' and '\dag' denote models where we used values for the hyper-parameters tuned by the authors. Specifically, '\ddag' refers to the QM9 dataset, while '\dag' refers to the ZINC dataset.}
\label{tab:results}
\begin{tabular}{ lcccccccccccc }
\toprule
& \multicolumn{6}{c}{\textbf{QM9 dataset}} &  \multicolumn{6}{c}{\textbf{ZINC dataset}}  \\
\cmidrule(lr{4pt}){2-7} \cmidrule(lr{4pt}){8-13}
\textbf{Model}      &\textbf{Rec.}    &\textbf{Val.}&\textbf{Nov.}&\textbf{Uniq.}   &\textbf{Div.}  &\textbf{Time}       &\textbf{Rec.}    &\textbf{Val.}&\textbf{Nov.}&\textbf{Uniq.}   &\textbf{Div.}  &\textbf{Time}              \\
\midrule
\mr{Char. VAE\textsuperscript{\dag}}    &49.9          &5.9             &92.2          &\mr{94.8}         &91.3              &\mr{02m}      &25.3              &0.9             &\dcb{}100        &\mr{91.4}          &98.2                &\mr{07m}    \\  
                                        &\(\pm\)50.0   &\(\pm\)23.5   &\(\pm\)26.8      &                   &\(\pm\)18.9       &            &\(\pm\)43.5       &\(\pm\)9.6      &\(\pm\)0         &                   &\(\pm\)7.0         &             \\ 
\mr{Gram. VAE\textsuperscript{\dag}}    &86.2          &12.6           &84.0            &\mr{59.3}          &98.7              &\mr{07m}      &55.8              &5.1             &\dcb{}100      &\mr{94.6}         &\dcb{}99.2        &\mr{21m}         \\
                                        &\(\pm\)34.5   &\(\pm\)33.2   &\(\pm\)36.7      &                   &\(\pm\)6.7        &            &\(\pm\)49.7       &\(\pm\)23.0     &\(\pm\)0       &                   &\(\pm\)4.5        &                \\
\mr{SD VAE\textsuperscript{\dag}}       &\dcb{}97.5    &16.0            &\dcb{}100      &\dcb{}             &\dcb{}99.6        &\mr{04m}    &\dcb{}77.4        &19.0            &\dcb{}100      &\dcb{}             &93.6              &\mr{24m}        \\
                                        &\(\pm\)16.0   &\(\pm\)36.7     &\(\pm\)0       &\dcb{}\mrm{100} &\(\pm\)1.1        &               &\(\pm\)41.8       &\(\pm\)39.2     &\(\pm\)0       &\dcb{}\mrm{100}    &\(\pm\)18.5       &                \\
\mr{Graph VAE\textsuperscript{\ddag}}   &13.6          &80.1          &45.6             &\mr{88.1}         &66.2              &\mr{01m}     &0.3               &62.6          &\dcb{}100      &\dcb{}               &71.5              &\mr{18m}        \\
                                        &\(\pm\)34.3   &\(\pm\)32.4   &\(\pm\)49.8      &                   &\(\pm\)28.0       &            &\(\pm\)4.6        &\(\pm\)48.4   &\(\pm\)0       &\dcb{}\mrm{100}      &\(\pm\)25.4       &                \\
\mr{Reg. GVAE\textsuperscript{\ddag}}   &7.3           &91.8          &49.8             &\mr{77.1}             &68.7            &\mr{01m}   &0.0               &86.5          &\dcb{}100      &\mr{90.3}         &97.9                  &\mr{19m}       \\
                                        &\(\pm\)26.0   &\(\pm\)27.5   &\(\pm\)50.0      &                   &\(\pm\)25.6       &            &\(\pm\)0.8        &\(\pm\)34.2   &\(\pm\)0       &                   &\(\pm\)7.0           &               \\
\mr{JT VAE\textsuperscript{\dag}}       &23.7          &99.9          &87.7             &\mr{89.5}         &60.9              &\mr{1.7h}    &50.2              &99.6          &99.9           &\mr{99.7}         &33.0                &\mr{55m}         \\
                                        &\(\pm\)42.5   &\(\pm\)2.7    &\(\pm\)32.8      &                   &\(\pm\)29.5       &            &\(\pm\)50.0       &\(\pm\)6.35    &\(\pm\)1.2     &                   &\(\pm\)21.78        &               \\
\mr{CGVAE\textsuperscript{\ddag, \dag}} &24.5          &\dcb{}100       &92.8          &\mr{98.3}         &76.1              &\mr{21m}       &0.4               &\dcb{}100   &\dcb{}100      &\mr{99.9}              &66.0              &\mr{15.5h}     \\
                                        &\(\pm\)27.9   &\(\pm\)0        &\(\pm\)19.1   &                   &\(\pm\)22.6       &              &\(\pm\)5.9        &\(\pm\)0       &\(\pm\)0       &                    &\(\pm\)22.8       &               \\
\mr{CCGVAE\textsuperscript{\ddag, \dag}}&55.4          &\dcb{}100       &88.5          &\mr{93.2}         &79.2              &\mr{1.5h}     &22.2               &\dcb{}100   &\dcb{}100     &\mr{92.8}                 &80.0               &\mr{21.5h}     \\
                                        &\(\pm\)49.7   &\(\pm\)0        &\(\pm\)31.9   &                   &\(\pm\)22.0       &             &\(\pm\)41.5        &\(\pm\)0       &\(\pm\)0       &                   &\(\pm\)16.7        &               \\
\mr{\textbf{RGCVAE (ours)}}             &94.8           &99.9          &96.4             &\mr{94}           &87.7              &\mr{01m}    &79.7               &99.9            &100       &\mr{61.1}          &98.9            &\mr{12m}                 \\            
                                        &\(\pm\)22.2    &\(\pm\)3.5       &\(\pm\)18.7    &                 &\(\pm\)19.2       &            &\(\pm\)40.2        &\(\pm\)1.2     &\(\pm\)0   &                   &\(\pm\)5.0     &                    \\
\bottomrule
\end{tabular}
\end{minipage}
\end{center}
\end{sidewaystable*}

    As explained in the main document, very often the models in the literature do not use all the same datasets and metrics considered in this work, and for this reason, it was necessary to reproduce the baseline results to calculate all the metrics. 
    For each model, we always used the official code with the hyper-parameters specified by the authors to reproduce the baselines' results.
    In Table \ref{tab:results} we have reported all the results obtained also in the datasets not originally considered by the authors. 
    Notice that we did not perform an extensive hyper-parameter search in these new datasets.

    \section{}
    \subsection{Evaluating Molecule Properties}
    We have performed additional evaluations on the presented RGCVAE.
    Specifically, we have explored the properties of the molecules generated by the model as done for~\cite{rigoni2020systematic, rigoni2020conditional} using some common metrics used in this area of research:
    \begin{enumerate}
        \item  \emph{Natural Product (NP)} which indicates how much the generated molecules structural space is similar to the one covered by natural products~\cite{ertl2008natural};
        \item \emph{Solubility (Sol.)} which indicates how much a molecule is soluble in water, an important property for drugs;
        \item \emph{Synthetic Accessibility Score (SAS)} which represents how easy~(0) or difficult~(100) it is to synthesize a molecule;
        \item \emph{Quantitative Estimation Drug-likeness (QED)} which indicates in percentage how likely it is that the molecule is a good candidate to become a drug.
    \end{enumerate}
    
    We report the results about our experiments in Table~\ref{tab:results_prop}.
    The table reports the evaluations of the properties and the number of epochs used for the training of each model. 
    Moreover, the last row reports the  average values of the properties for each dataset. 
    
    \begin{table*}
\small
\begin{center}
\begin{minipage}{0.94\linewidth}
\centering
\caption{Average and standard deviation (where applicable) of different evaluation metrics on the QM9 and ZINC data-sets. The symbols '\ddag' and '\dag' denote models where we used values for the hyper-parameters tuned by the authors. Specifically, '\ddag' refers to the QM9 dataset, while '\dag' refers to the ZINC dataset.}
\label{tab:results_prop}
\begin{tabular}{ lcccccc}
\toprule
\textbf{Model trained on QM9}    &$\uparrow$\textbf{\%NP}  &$\uparrow$\textbf{\%Sol.} &$\downarrow$\textbf{\%SAS} &$\uparrow$\textbf{\%QED} &\textbf{T.*Epoch}&\textbf{N.Epochs}    \\
\midrule
\mr{Character VAE}                              &88.79          &\dcb{}46.55    &29.10          &30.02           &\mr{00h:02m}      &\mr{100}   \\
                                                &\(\pm\)11.74   &\(\pm\)32.71   &\(\pm\)28.52   &\(\pm\)19.55    &                  &           \\
\mr{Grammar VAE}                                &83.34          &35.85          &52.3          &35.18            &\mr{00h:07m}      &\mr{100}   \\
                                                &\(\pm\)15.45   &\(\pm\)19.46   &\(\pm\)31.63   &\(\pm\)11.53    &                  &           \\
\mr{Syntax Directed VAE}                        &88.89          &26.2           &14.65          &31.37           &\mr{00h:04m}      &\mr{500}   \\
                                                &\(\pm\)10.64   &\(\pm\)22.26   &\(\pm\)35.15   &\(\pm\)11.18    &                  &           \\
\mr{Graph VAE\textsuperscript{\ddag}}           &94.71          &35.92          &29.72          &48.25           &\mr{00h:01m}      &\mr{200}   \\
                                                &\(\pm\)10.82   &\(\pm\)13.49   &\(\pm\)28.27   &\(\pm\)9.53     &                  &           \\
\mr{Regularized GVAE\textsuperscript{\ddag}}    &95.77          &39.38          &30.58          &\dcb{}48.79     &\mr{00h:01m}      &\mr{150}   \\
                                                &\(\pm\)9.26    &\(\pm\)14.52   &\(\pm\)24.69   &\(\pm\)7.83     &                  &           \\
\mr{Junction Tree VAE}                          &90.77          &27.25          &19.62          &46.89           &\mr{01h:40m}      &\mr{10}    \\
                                                &\(\pm\)16.00   &\(\pm\)13.17   &\(\pm\)21.18   &\(\pm\)7.73     &                  &           \\
\mr{CGVAE\textsuperscript{\ddag}}               &93.80          &28.62          &\dcb{}10.28    &47.91           &\mr{00h:21m}      &\mr{10}    \\
                                                &\(\pm\)5.62   &\(\pm\)12.38    &\(\pm\)16.11    &\(\pm\)7.04    &                  &           \\
\mr{CCGVAE\textsuperscript{\ddag}}              &\dcb{}96.13    &35.58          &17.08          &46.62           &\mr{01h:30m}      &\mr{10}    \\
                                                &\(\pm\)8.64    &\(\pm\)11.91    &\(\pm\)22.96    &\(\pm\)7.51   &                  &           \\
\mr{\textbf{RGCVAE (ours)}}                     &86.93          &25.4          &11.17          &40.92         &\mr{00h:01m}       &\mr{200}   \\
                                                &\(\pm\)13.68    &\(\pm\)14.03  &\(\pm\)20.34   &\(\pm\)9.55   &                    &           \\
\midrule
                                         & 88.52         & 27.91         &21.86          &46.12           &                  &           \\
\mrm{QM9 Properties' Scores}             &\(\pm\)17.75   &\(\pm\)13.76   &\(\pm\)22.88   &\(\pm\)7.76     &                  &           \\
\bottomrule
\end{tabular}
\\\vspace*{0.1cm}
\begin{tabular}{ lcccccc}
\toprule
\textbf{Model trained on ZINC}    &$\uparrow$\textbf{\%NP}  &$\uparrow$\textbf{\%Sol.}&$\downarrow$\textbf{\%SAS} &$\uparrow$\textbf{\%QED} &\textbf{T.*Epoch} &\textbf{N.Epochs}    \\
\midrule
\mr{Character VAE\textsuperscript{\dag}}        &80.82          &29.60          &31.11              &38.70          &\mr{00h:07m}   &\mr{100}   \\
                                                &\(\pm\)12.83   &\(\pm\)17.60   &\(\pm\)30.14       &\(\pm\)10.63   &               &           \\
\mr{Grammar VAE\textsuperscript{\dag}}          &80.99          &50.24          &26.75              &25.42          &\mr{00h:21m}   &\mr{100}   \\
                                                &\(\pm\)11.40   &\(\pm\)33.65   &\(\pm\)33.14       &\(\pm\)14.91   &               &           \\
\mr{Syntax Directed VAE\textsuperscript{\dag}}  &77.84          &55.94          &\dcb{}14.46        &39.45          &\mr{00h:24m}   &\mr{500}   \\
                                                &\(\pm\)19.76   &\(\pm\)27.51   &\(\pm\)24.14       &\(\pm\)20.98   &               &           \\ 
\mr{Graph VAE}                                  &90.68          &80.79          &28.07              &45.96          &\mr{00h:18m}   &\mr{400}   \\
                                                &\(\pm\)11.71   &\(\pm\)17.33   &\(\pm\)20.14       &\(\pm\)18.69   &               &           \\ 
\mr{Regularized GVAE}                           &\dcb{}95.88    &\dcb{}94.42    &44.64              &34.41          &\mr{00h:19m}   &\mr{300}   \\
                                                &\(\pm\)6.84    &\(\pm\)9.61    &\(\pm\)25.14       &\(\pm\)13.26   &               &           \\
\mr{Junction Tree VAE\textsuperscript{\dag}}    &52.20          &48.06          &44.74              &\dcb{}75.05    &\mr{07h:55m}   &\mr{10}    \\
                                                &\(\pm\)17.12   &\(\pm\)18.48   &\(\pm\)24.39       &\(\pm\)13.40   &               &           \\
\mr{CGVAE\textsuperscript{\dag}}                &81.38          &57.76          &16.25              &65.14          &\mr{15h:30m}   &\mr{3}     \\
                                                &\(\pm\)15.98   &\(\pm\)20.04   &\(\pm\)21.63       &\(\pm\)16.39   &               &           \\
\mr{CCGVAE\textsuperscript{\dag}}               &94.28          &63.54          &19.95              &52.41          &\mr{21h:30m}  &\mr{3}     \\
                                                 &\(\pm\)10.08   &\(\pm\)20.11   &\(\pm\)23.10       &\(\pm\)16.52   &               &           \\
\mr{\textbf{RGCVAE (ours)}}                     &86.62          &40.74          &36.59              &43.01          &\mr{00h:12m}   &\mr{200}   \\
                                                &\(\pm\)13.13    &\(\pm\)12.37  &\(\pm\)32.22       &\(\pm\)7.56     &              &           \\
\midrule
                                                 & 42.08         & 56.11         & 55.95             & 73.18         &               &           \\
\mrm{ZINC Properties' Scores}                    &\(\pm\)18.37   &\(\pm\)17.44   &\(\pm\)22.90       &\(\pm\)13.86   &               &           \\
\bottomrule
\end{tabular}
\end{minipage}
\end{center}
\end{table*}

    On the QM9 dataset, the properties evaluation performed on the  molecules generated by RGCVAE returned results close to those calculated on the dataset, supporting again the fact that our model does learn well the probably distributions of the input dataset molecules. 
    When considering the ZINC dataset, RGCVAE still shows good values according to those calculated on the dataset. \emph{NP} and  \emph{QED} are far from the values exhibited by the molecule in the dataset, probably because of the high \emph{Diversity} of its generated molecules (see Table 1 in the main paper). Since our model generates structures very dissimilar from the ones in training, the generated molecules are dissimilar also in their properties. This behavior can be adjusted reducing the expressiveness of the model. We will explore this aspect in future works.

    Table~\ref{tab:ablation_prop} follows the format of  Table ~\ref{tab:results_prop}, 
    reporting the property metrics results obtained in the ablation study, using either representation 1 or 3.
    
    \begin{sidewaystable*}
\begin{center}
\begin{minipage}{\linewidth}
\caption{This table presents the ablation study results obtained by RGCVAE model. Notation and format are the same used in Table~\ref{tab:results_prop}} 
\label{tab:ablation_prop}
\begin{tabular}{ cccccccccccc}
\toprule
\multicolumn{3}{c}{}&\multicolumn{4}{c}{\textbf{QM9}} & \multicolumn{4}{c}{\textbf{ZINC}}  \\
\cmidrule(lr{4pt}){4-7} \cmidrule(lr{4pt}){8-11}
\textbf{Atom Rep.}& \textbf{Hist.}& \textbf{Encoding Network}  &$\uparrow$\textbf{\%NP}  &$\uparrow$\textbf{\%Sol.} &$\downarrow$\textbf{\%SAS} &$\uparrow$\textbf{\%QED}&$\uparrow$\textbf{\%NP}  &$\uparrow$\textbf{\%Sol.} &$\downarrow$\textbf{\%SAS} &$\uparrow$\textbf{\%QED}      \\
\midrule
\mr{1}      &\mr{\XSolidBold}      &\mr{RGIN}  &94.09          &34.66          &28.48          &46.93          &78.43          &32.77          &51.89          &37.41            \\
            &                       &           &\(\pm\)10.49   &\(\pm\)13.31   &\(\pm\)28.58   &\(\pm\)7.07    &\(\pm\)7.87    &\(\pm\)8.54   &\(\pm\)27.15   &\(\pm\)2.39      \\                     
\mr{1}      &\mr{\CheckmarkBold}    &\mr{RGIN}  &85.07          &29.57          &17.65          &42.36          &93.05          &42.69          &21.15          &46.93            \\
            &                       &           &\(\pm\)14.57    &\(\pm\)12.78   &\(\pm\)23.61   &\(\pm\)9.66    &\(\pm\)10.77   &\(\pm\)14.56   &\(\pm\)25.66   &\(\pm\)9.93      \\          
\mr{3}      &\mr{\XSolidBold}       &\mr{RGIN}  &94.30          &31.48          &25.13          &45.25          &78.27          &33.0          &49.78          &37.21            \\
            &                       &           &\(\pm\)9.63     &\(\pm\)14.87   &\(\pm\)28.88   &\(\pm\)7.75    &\(\pm\)8.01    &\(\pm\)8.65    &\(\pm\)28.21   &\(\pm\)2.43      \\  
\mr{3}      &\mr{\CheckmarkBold}    &\mr{RGIN}  &86.93          &25.40          &11.17          &40.92          &86.62          &40.74          &36.59          &43.01            \\
            &                       &           &\(\pm\)3.53    &\(\pm\)14.03   &\(\pm\)20.34   &\(\pm\)9.55     &\(\pm\)13.13   &\(\pm\)12.37   &\(\pm\)32.33   &\(\pm\)7.56      \\
\mr{3}      &\mr{\CheckmarkBold}    &\mr{GIN}   &91.48          &26.77          &16.25          &41.77          &83.6          &41.21          &36.91          &42.46            \\
            &                       &           &\(\pm\)11.32   &\(\pm\)13.81   &\(\pm\)24.26   &\(\pm\)8.49    &\(\pm\)11.46   &\(\pm\)10.92   &\(\pm\)34.96   &\(\pm\)6.98      \\
\mr{3}      &\mr{\CheckmarkBold}    &\mr{RGCNC} &85.30          &29.16          &16.85          &41.28          &90.54          &41.04          &26.01          &45.24            \\
            &                       &           &\(\pm\)14.31   &\(\pm\)13.60   &\(\pm\)23.26   &\(\pm\)9.60    &\(\pm\)11.37   &\(\pm\)14.27   &\(\pm\)29.85   &\(\pm\)9.68      \\
\midrule
\multicolumn{3}{c}{}                           &88.52          &27.91          &21.86          &46.12          &42.08         &56.11           &55.95          &73.18                          \\
\multicolumn{3}{c}{\mrm{Properties' Scores}}        &\(\pm\)17.75   &\(\pm\)13.76   &\(\pm\)22.88   &\(\pm\)7.76    &\(\pm\)18.37   &\(\pm\)17.44   &\(\pm\)22.90       &\(\pm\)13.86               \\
\bottomrule
\end{tabular}
\end{minipage}
\end{center}
\end{sidewaystable*}
    For the sake of interpretability,  we report in the first column of the table the numerical ids corresponding to the three atoms representations:
    \begin{enumerate}
        \item atom type, e.g. ``\textbf{C}'' for  carbon;
        \item atom type, total valence and formal charge, e.g. ``\textbf{C4(0)}'' for carbon atom with total valence 4 and formal charge 0.
        \item atom type, total valence, formal charge, presence of chiral property e.g. ``\textbf{O3(1)0}'' for an oxygen atom with total valence 3, formal charge 1 and without the chiral property. 
    \end{enumerate}
    It is interesting to notice that the use of the histogram of valences drives the generation of molecules towards those that are simpler to synthesize (lower SAS) and it increases Reconstruction, Novelty and Uniqueness (see Table 3 in the main paper). However, it seems that the use of histograms slightly decrease QED values.

    \begin{table*}
\small
\begin{center}
\begin{minipage}{\linewidth}
\caption{This table presents the ablation study results obtained by RGCVAE model. Notation and format are the same used in Table~\ref{tab:results_prop}} 
\label{tab:ablation_lambda_prop}
\begin{tabular}{ ccccccccc }
\toprule
&\multicolumn{4}{c}{\textbf{QM9}} & \multicolumn{4}{c}{\textbf{ZINC}} \\
\cmidrule(lr{4pt}){2-5} \cmidrule(lr{4pt}){6-9}
$\bm{\lambda_1}$&$\uparrow$\textbf{\%NP}  &$\uparrow$\textbf{\%Sol.} &$\downarrow$\textbf{\%SAS} &$\uparrow$\textbf{\%QED}&$\uparrow$\textbf{\%NP}  &$\uparrow$\textbf{\%Sol.} &$\downarrow$\textbf{\%SAS} &$\uparrow$\textbf{\%QED}    \\
\midrule
\mr{0.01} &91.47          &28.58          &13.07          &42.74          &86.62          &40.74          &36.59          &43.01           \\
          &\(\pm\)11.87   &\(\pm\)13.77   &\(\pm\)21.33   &\(\pm\)8.94    &\(\pm\)13.13    &\(\pm\)12.37   &\(\pm\)32.22   &\(\pm\)7.56       \\
\mr{0.05} &86.93          &25.04          &11.17          &40.92          &92.72          &44.53          &6.16          &43.22           \\
          &\(\pm\)13.68    &\(\pm\)14.03   &\(\pm\)20.34   &\(\pm\)9.55   &\(\pm\)9.21   &\(\pm\)18.12   &\(\pm\)14.83   &\(\pm\)13.43       \\
\mr{0.1}  &84.27          &25.75          &11.77          &40.12          &87.51          &45.79          &18.69          &42.36             \\
          &\(\pm\)14.1     &\(\pm\)13.67   &\(\pm\)20.02   &\(\pm\)10.08    &\(\pm\)11.72    &\(\pm\)16.13    &\(\pm\)26.78   &\(\pm\)10.92      \\
\mr{0.2}  &78.67          &29.49          &17.85          &40.57            &88.68          &47.82          &4.03          &39.82           \\
          &\(\pm\)14.87    &\(\pm\)14.19   &\(\pm\)23.09   &\(\pm\)10.08    &\(\pm\)10.2   &\(\pm\)19.31   &\(\pm\)12.79   &\(\pm\)14.69      \\
\midrule
      &88.52          &27.91          &21.86          &46.12          &42.08         &56.11           &55.95          &73.18          \\
\mrm{Properties' Scores} &\(\pm\)17.75   &\(\pm\)13.76   &\(\pm\)22.88   &\(\pm\)7.76    &\(\pm\)18.37   &\(\pm\)17.44   &\(\pm\)22.90       &\(\pm\)13.86          \\
\bottomrule
\end{tabular}
\end{minipage}
\end{center}
\end{table*}
    We report the molecules' properties results of our model as the $\lambda_1$ hyper-parameter values changes in Table \ref{tab:ablation_lambda_prop}.
    From these results, we can see that in both QM9 and ZINC datasets, the \emph{QED} values tends to decrease as the $\lambda_1$ hyper-parameter values increase.

    \section{}
    \subsection{Implementation Details}
    \label{app:impl_details}
    In the following,  we report the implementation details of our model.
    When it is not explicitly indicated, all the neural networks use the \emph{leaky ReLU} activation function.
    
    \subsection{Encoder}
    
    In our implemented model we used the following values for the encoder parameters: $K=5$, $s_h=70$, $s_{lt}=70$, \mbox{$d_p=1$}, $\epsilon^{(k)} = 0\ \forall k \in [1, K]$. 
    
    Moreover, $\text{MLP}^{(k)}_{\Phi_e(u,v)}$ is implemented as a linear layer followed by \emph{leaky ReLU}, $\text{MLP}^{(k)}$ is a multi-layer perceptron\footnote{It preserves the input dimension.} (MLP) with only one hidden layer and  batch normalization before the \emph{leaky ReLU} activation, while $\text{MLP}_\mu$ and $\text{MLP}_\Sigma$ are feed-forward neural networks with \emph{leaky ReLU} activation. In particular, $\text{MLP}_\Sigma$ clamps the activation to be at maximum 2.5.
    
    \subsection{Decoder}
    
    As explained in the main paper,  we adopted the procedure provided in~\cite{rigoni2020conditional},  where the atom type assignment process is conditioned by the previously assigned atom types. 
    Let $\balpha^u_0$ be the histogram where all the valences are 0, and \mbox{$t \in \{1, \ldots, m\}$}, then each atom type $\tau_{t}$ is predicted according to the following equations:
    \begin{align}
    \balpha^d_{t} &= \balpha_{t-1} - \balpha^u_{t-1}, \\
    \be_{t} &= K(\bz_t, \balpha^d_{t}, \balpha^u_{t-1}), \\
    \bbr_{t} &= \text{Concat}(\bz_t, \be_t), \\
    \tau_{t} &= \text{Sample}_\text{type}(F(\bbr_{t}), \balpha^d_{t}), \\
    \balpha^u_{t} &= \text{Update}(\tau_{t}, \balpha^u_{t-1}), \\
    \balpha_{t} &= \text{Sample}_\text{distr}({\cal H}, \balpha^u_t), 
    \end{align}
    where
    $\balpha^d_t$ is the difference histogram,
    $\balpha^u_t$ is the updated histogram,
    $K(\bz_t, \balpha^d_t, \balpha^u_t)$ is a function that maps the inputs to a new representation $\be_t$,
    $\text{Concat}(\bz_t, \be_t)$ is the concatenation function,
    $F(\bbr_t)$ is a function that computes a probability distribution on the atom types,
    $\text{Sample}_\text{type}(F(\bbr_t), \balpha^d_t)$ samples the atom type from the probabilities computed by $F$ masking all the atoms whose valences have a zero-value in the histogram $\alpha^d_t$.
    $\text{Update}(\tau_t, \balpha^u_t)$ is a function that updates the histogram $\balpha^u_t$ with the valence of the sampled 
    $\text{Sample}_\text{distr}({\cal H}, \balpha^u_t)$: at training time this function returns the histogram $\balpha^v_0$. 
    At generation time, it samples from ${\cal H}$ a new histogram $\balpha_{t+1}$ with at least $m$ atoms, such that $\balpha^u_t$ is compatible with $\balpha_{t+1}$.
    Notice that in our implementation, in generation we have kept the same training behaviour for the function $\text{Sample}_\text{distr}({\cal H}, \balpha^u_t)$.
    
    In our atom decoding implementation, the function $K$ is a linear transformation followed by a \emph{tanh} activation function, $F$ is an MLP with one hidden layer of the same size of the input and \emph{leaky ReLU} as activation function, and $S_n=120$. 
    
    We apply a batch normalization layer to all the nodes representation at the end of the atoms decoding.
    
    Regarding the edge decoding implementation, we use the following parameter values: $dim(s_v)=190$ and \mbox{$dim(\bphi_{(u,v)})=570$}.
    $C$ and  $L$ are implemented as two MLP networks with two hidden layers of dimension $590$ and $190$, respectively, and \emph{leaky ReLU} as activation function. In particular, $C$ uses the \emph{sigmoid} activation as last activation function, while $L$ uses \emph{softmax} activation function in output.
    
    \subsection{Optimization}
    
    The optimization function $F$ is implemented as a feed-forward neural network without hidden layers and with \emph{Leaky\_ReLU} activation function. 
    $Q^1_p$ and $Q^2_p$ are both feed-forward networks without hidden layers for each $p \in \Theta_p$.
    Regarding the optimization search on ZINC, we set the gradient ascent step to $0.01$, we always fix the histogram $\alpha_0$ for each node decoding iteration  and we use the $softmax$ function to predict the type of edges and nodes.

    \section{}
    \subsection{Hyper-parameter Optimization}
    
    We performed a grid search on the $\lambda_1$ hyper-parameter with the following values: $\{0.2, 0.1, 0.05, 0.01\}$ for both QM9, and ZINC datasets. Specifically, we have noticed that the \emph{reconstruction} performance of the model increases with low $\lambda_1$ values, but at the same time, \emph{uniqueness} and \emph{novelty} decrease. The opposite behavior has been observed for values of $\lambda_1$ close to $1$.
    
    We did not tune the hyper-parameter $\lambda_2$ that we fix at 10, while we selected the  learning rate from the following set of values: $\{0.005, 0.001, 0.0005\}$. The batch size is fixed at $100$ for both  datasets, while the maximum training epochs are $200$ for ZINC and $300$ for QM9, respectively.
    We have obtained the best performance in the validation set using $\lambda_1=0.05$, $\lambda_2=10$, and $0.001$ as learning rate for QM9, and $\lambda_1=0.001$, $\lambda_2=10$, and $0.001$ as learning rate for ZINC.
    
    During the generation of new molecules, we sample the atoms and bonds according to the predicted probabilities, while during reconstruction of an input molecule, we have used the $argmax$ function on the returned probabilities.

    \subsection{Qualitative Examples}
\begin{figure*}[t]
    \centering
    \includegraphics[width=0.8\textwidth]{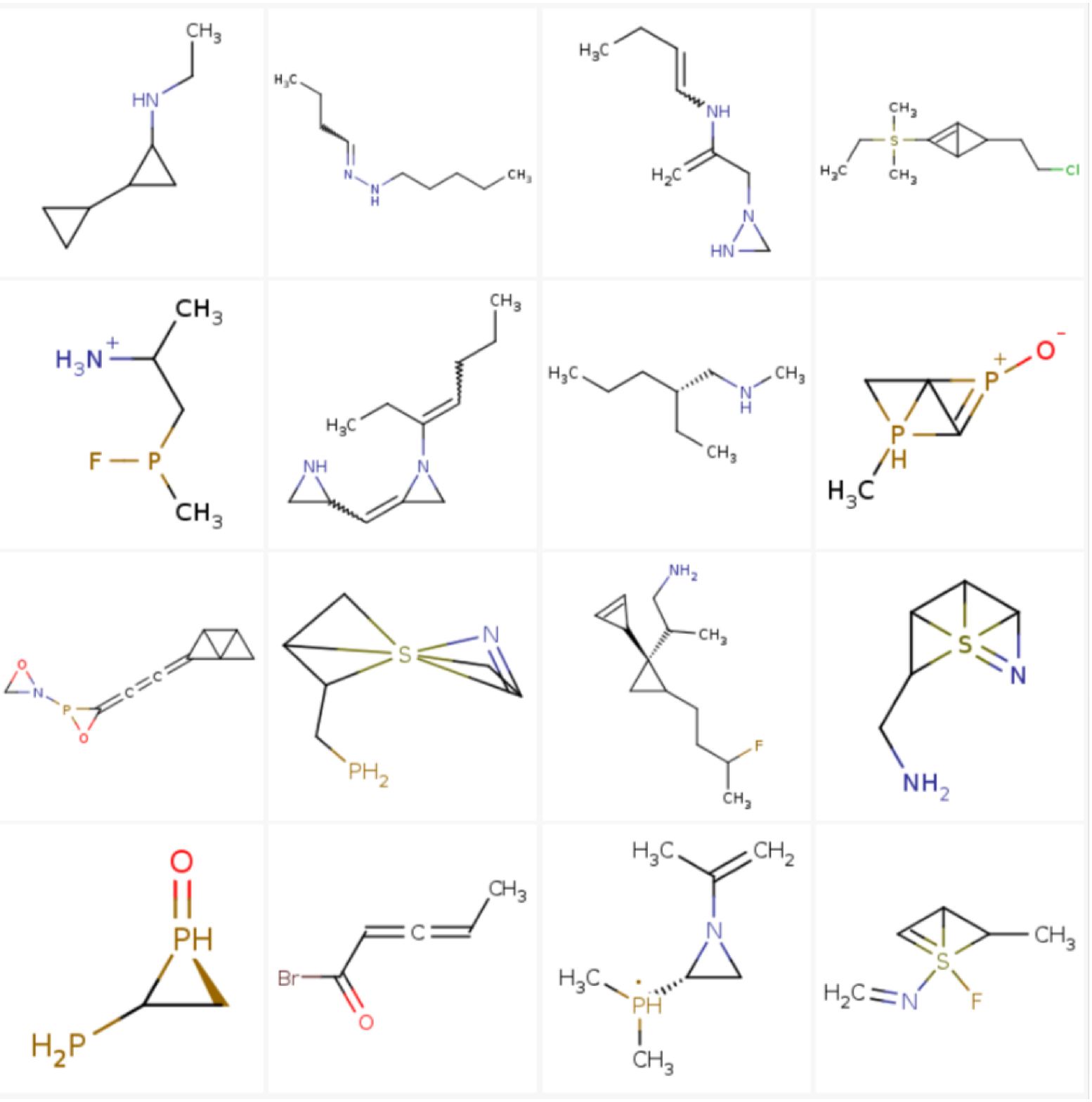}
    \caption{\label{fig:examples} Examples of generated molecules.}
    \end{figure*}

    We report in Fig.~\ref{fig:examples} some qualitative examples obtained in generation on the ZINC dataset.
    
\end{appendices}

\end{document}